\documentclass{tlp}
\usepackage{aopmath}
\usepackage{algorithm}
\usepackage{amssymb}
\usepackage{algpseudocode}
\usepackage{tikz}
\usepackage{multicol}
\usepackage{subfig}
\usepackage{listings}
\usepackage{url}
\usetikzlibrary{arrows.meta}
\newtheorem{definition}{Definition} 
\newtheorem{example}{Example} 
\newcounter{theoremPlaceHolder}
\usepackage{fancyvrb}

\providecommand{\codeif}{\texttt{:- }}
\providecommand{\ruleend}{\texttt{.}}

\providecommand{\naf}[1]{\texttt{not }#1}
\providecommand{\asp}[1]{\mbox{$\mathtt{#1}$}}

\providecommand{\lasne}{\textit{ILP}_{\textit{\scriptsize LAS}}^{\textit{ \scriptsize noise}}}
\providecommand{\loasne}{\textit{ILP}_{\textit{\scriptsize LOAS}}^{\textit{ \scriptsize noise}}}
\providecommand{\wrt}{w.r.t.\ }

\newcommand{\proofindent}[1]{

  \begingroup
  \setlength{\parindent}{0pt}

  \addtolength{\leftskip}{5mm}

  #1

  \endgroup

}

\edef\appendixproofs{}

\newcommand{\addtoappendixproofs}[1]{
  \expandafter\def\expandafter\appendixproofs\expandafter{\appendixproofs #1}
}

\newcommand{\appendixTheorem}[3]{
  \begin{theorem}\label{thm:#1}
    #2
  \end{theorem}

  \addtoappendixproofs{
    \setcounter{theoremPlaceHolder}{\value{theorem}}
    \setcounter{theorem}{\ref{thm:#1}}
    \addtocounter{theorem}{-1}
    \begin{theorem}
      #2
    \end{theorem}
    \setcounter{theorem}{\value{theoremPlaceHolder}}
    \begin{proof}
      #3
    \end{proof}
  }
}

\begin{document}

\title{Conflict-driven Inductive Logic Programming}

\author[M. Law]{Mark Law\\ ILASP Limited, UK\\ \email{mark@ilasp.com}}


\maketitle

\begin{abstract}
  The goal of Inductive Logic Programming (ILP) is to learn a program that
  explains a set of examples. Until recently, most research on ILP targeted
  learning Prolog programs. The ILASP system instead learns Answer Set Programs
  (ASP). Learning such expressive programs widens the applicability of ILP
  considerably; for example, enabling preference learning, learning common-sense
  knowledge, including defaults and exceptions, and learning non-deterministic
  theories.

  Early versions of ILASP can be considered \emph{meta-level} ILP approaches,
  which encode a learning task as a logic program and delegate the search to an
  ASP solver. More recently, ILASP has shifted towards a new method, inspired
  by conflict-driven SAT and ASP solvers. The fundamental idea of the approach,
  called \emph{Conflict-driven ILP} (CDILP), is to iteratively interleave the
  search for a hypothesis with the generation of constraints which explain why
  the current hypothesis does not cover a particular example. These
  \emph{coverage constraints} allow ILASP to rule out not just the current
  hypothesis, but an entire class of hypotheses that do not satisfy the
  coverage constraint.

  This paper formalises the CDILP approach and presents the ILASP3 and ILASP4
  systems for CDILP, which are demonstrated to be more scalable than previous
  ILASP systems, particularly in the presence of noise.

  \textbf{Under consideration in Theory and Practice of Logic Programming (TPLP).}
\end{abstract}


\section{Introduction}
\label{sec:intro}
Inductive Logic Programming (ILP)~\cite{Muggleton1991} systems aim to find a
set of logical rules, called a hypothesis, that, together with some existing
background knowledge, explain a set of examples. Unlike most ILP systems, which
usually aim to learn Prolog programs, the ILASP (Inductive Learning of Answer
Set Programs) systems~\cite{JELIAILASP,ILASP_thesis,ALP2020} can learn Answer
Set Programs (ASP), including normal rules, choice rules, disjunctive rules,
and hard and weak constraints. ILASP's learning framework has been proven to
generalise existing frameworks and systems for learning ASP
programs~\cite{AIJ17}, such as the brave learning framework~\cite{Sakama2009},
adopted by almost all previous systems (e.g.\ XHAIL~\cite{ray2009nonmonotonic},
ASPAL~\cite{corapi2012}, ILED~\cite{ILED}, RASPAL~\cite{raspal}), and the less
common cautious learning framework~\cite{Sakama2009}.
Brave systems require the examples to be covered in at least one answer set of
the learned program, whereas cautious systems find a program which covers the
examples in every answer set.
The results in~\cite{AIJ17} show that some ASP programs cannot be learned with
either a brave or a cautious approach, and that to learn ASP programs in
general, a combination of both brave and cautious reasoning is required.
ILASP's learning framework enables this combination, and is capable of learning
the full class of ASP programs~\cite{AIJ17}.
ILASP's generality has allowed it to be applied to a wide range of
applications, including event detection~\cite{ACS18}, preference
learning~\cite{ICLP15}, natural language
understanding~\cite{chabierski2017machine}, learning game
rules~\cite{cropper2019inductive}, grammar induction~\cite{law2019representing}
and automata induction~\cite{danielf,danielf2}.

Throughout the last few decades, ILP systems have evolved from early
bottom-up/top-down learners, such
as~\cite{quinlan1990learning,muggleton1995inverse,srinivasan2001aleph}, to more
modern systems, such as~\cite{metagol,hexmil,corapi2010,corapi2012}, which take
advantage of logic programming systems to solve the task. These recent ILP
systems, commonly referred to as \emph{meta-level} systems, work by
transforming an ILP learning problem into a meta-level logic program whose
solutions can be mapped back to the solutions of the original ILP problem.  The
specific notion of solution of the meta-level logic program differs from system
to system, but to give an example, in ASPAL a brave induction task is mapped
into an ASP program whose answer sets each encode a brave inductive solution of
the original task. When compared to older style bottom-up/top-down learners,
the advantage of meta-level approaches is that they tend to be complete (which
means that most systems are guaranteed to find a solution if one exists) and
because they use off-the-shelf logic programming systems to perform the search
for solutions, problems which had previously been difficult challenges for ILP
(such as recursion and non-observational predicate learning) are much simpler.
Unlike traditional ILP approaches, which incrementally construct a hypothesis
based on a single seed example at a time, meta-level ILP systems tend to be
\emph{batch learners} and consider all examples at once.  This can mean that
they lack scalability on datasets with large numbers of examples.

At first glance, the earliest ILASP systems (ILASP1~\cite{JELIAILASP} and
ILASP2~\cite{ICLP15}) may seem to be meta-level systems, and they do indeed
involve encoding a learning task as a meta-level ASP program; however, they are
actually in a more complicated category. Unlike ``pure'' meta-level systems,
the ASP solver is not invoked on a fixed program, and is instead (through the
use of multi-shot solving~\cite{clingo5}) incrementally invoked on a program
that is growing throughout the execution.
With each new version, ILASP has shifted further away from pure meta-level
approaches, towards a new category of ILP system, which we call
\emph{conflict-driven}. Conflict-driven ILP systems, inspired by
conflict-driven SAT and ASP solvers, iteratively construct a set of constraints
on the solution space that must\footnote{In the case of noisy examples, these
are ``soft'' constraints that should be satisfied, but can be ignored for a
penalty.} be satisfied by any inductive solution. In each iteration, the solver
finds a program $H$ that satisfies the current constraints, then searches for a
\emph{conflict} $C$, which corresponds to a reason why $H$ is not an (optimal)
inductive solution.  If none exists, then $H$ is returned; otherwise, $C$ is
converted to a new \emph{coverage constraint} which the next programs must
satisfy. The process of converting a conflict into a new coverage constraint is
called \emph{conflict analysis}.

This paper formalises the notion of Conflict-driven Inductive Logic Programming
(CDILP) which is at the core of the most recent two ILASP systems
(ILASP3~\cite{ILASP_thesis} and ILASP4). Although ILASP3 was released in 2017
and previously presented in Mark Law's PhD thesis~\cite{ILASP_thesis}, and has
been evaluated on several applications~\cite{ACS18}, the approach has not been
formally published until now. In fact, despite being equivalent to the
formalisation in this paper, the definition of ILASP3 in~\cite{ILASP_thesis}
uses very different terminology.

This paper first presents CDILP at an abstract level and proves that, assuming
certain guarantees are met, any instantiation of CDILP is guaranteed to find an
optimal solution for any learning task, provided at least one solution exists.
ILASP3 and ILASP4 are both instances of the CDILP aproach, with the difference
being their respective methods for performing conflict analysis. These are
formalised in Section 4, with a discussion of the strengths and weaknesses of
each method. In particular, we identify one type of learning task on which
ILASP4 is likely to significantly outperform ILASP3.

CDILP is shown through an evaluation to be significantly faster than previous
ILASP systems on tasks with noisy examples. One of the major advantages of the
CDILP approach is that it allows for \emph{constraint propagation}, where a
coverage constraint computed for one example is \emph{propagated} to another
example. This means that the conflict analysis performed on one example does
not need to be repeated for other similar examples, thus improving efficiency.

The CDILP approach (as ILASP3) has already been evaluated~\cite{ACS18} on
several real datasets and compared with other state-of-the-art ILP systems.
Unlike ILASP, these systems do not guarantee finding an \emph{optimal} solution
of the learning task (in terms of the length of the hypothesis, and the
penalties paid for not covering examples). ILASP finds solutions which are on
average better quality than those found by the other systems (in terms of the
$F_1$-score on a test set of examples)~\cite{ACS18}. The evaluation in this
paper compares the performance of ILASP4 to ILASP3 on several synthetic
learning tasks, and shows that ILASP4 is often significantly faster than
ILASP3.

The CDILP framework is entirely modular, meaning that users of the ILASP system
can replace any part of the CDILP approach with their own method; for instance,
they could define a new method for conflict analysis or constraint propagation
to increase performance in their domain. Providing their new method shares the
same correctness properties as the original modules in ILASP, their customised
CDILP approach will still be guaranteed to terminate and find an optimal
solution. This customisation is supported in the ILASP implementation through
the use of a new Python interface (called PyLASP).

The rest of the paper is structured as follows. Section 2 recalls the necessary
background material. Section 3 formalises the notion of Conflict-driven ILP.
Section 4 presents several approaches to conflict analysis. Section 5 gives an
evaluation of the approach. Finally, Sections 6 and 7 present the related work
and conclude the paper.

\section{Background}
\label{sec:background}

This section introduces the background material that is required to understand
the rest of the paper. First, the fundamental Answer Set Programming concepts
are recalled, and then the \emph{learning from answer sets} framework used by
the ILASP systems is formalised.

\subsection{Answer Set Programming}

A \emph{disjunctive rule} $R$ is of the form $\asp{h_1\lor\ldots\lor h_m\codeif
b_1,\ldots,b_n,}$ $\asp{\naf c_1,\ldots,\naf c_o}$, where $\lbrace
\asp{h_1},\ldots,\asp{h_m} \rbrace$, $\lbrace\asp{b_1},\ldots,\asp{b_n}\rbrace$
and $\lbrace\asp{c_1},\ldots,\asp{c_o}\rbrace$ are sets of atoms denoted
$\textit{head}(R)$, $\textit{body}^{+}(R)$ and $\textit{body}^{-}(R)$,
respectively. A \emph{normal rule} $R$ is a disjunctive rule such that
$|\textit{head}(R)| = 1$. A \emph{definite rule} $R$ is a disjunctive rule such
that $|\textit{head}(R)| = 1$ and $|\textit{body}^{-}(R)| = 0$. A \emph{hard
constraint} $R$ is a disjunctive rule such that $|\textit{head}(R)| = 0$. Sets
of disjunctive, normal and definite rules are called disjunctive, normal and
definite logic programs (respectively).

Given a (First-order) disjunctive logic program $P$, the Herbrand base
($\textit{HB}_{P}$) is the set of all atoms constructed using constants,
functions and predicates in $P$. The program $\textit{ground}(P)$ is
constructed by replacing each rule with its ground instances (using only atoms
from the Herbrand base). An (Herbrand) interpretation $I$ (of $P$) assigns each
element of $\textit{HB}_{P}$ to $\asp{\top}$ or $\asp{\bot}$, and is usually
written as the set of all elements in $\textit{HB}_{P}$ that $I$ assigns to
$\asp{\top}$. An interpretation $I$ is a model of $P$ if it satisfies every
rule in $P$; i.e.\ for each rule $R \in \textit{ground}(P)$ if
$\textit{body}^{+}(R)\subseteq I$ and $\textit{body}^{-}(R)\cap I = \emptyset$
then $\textit{head}(R)\cap I \neq \emptyset$. A model of $P$ is \emph{minimal}
if no strict subset of $P$ is also a model of $P$. The \emph{reduct} of $P$ wrt
$I$ (denoted $P^{I}$) is the program constructed from $\textit{ground}(P)$ by
first removing all rules $R$ such that $\textit{body}^{-}(R)\cap I \neq
\emptyset$ and then removing all remaining negative body literals from the
program. The \emph{answer sets} of $P$ are the interpretations $I$ such that
$I$ is a minimal model of $P^I$. The set of all answer sets of $P$ is denoted
$\textit{AS}(P)$.

There is another way of characterising answer sets, by using \emph{unfounded
subsets}. Let $P$ be a disjunctive logic program and $I$ be an interpretation.
A subset $U\subseteq I$ is unfounded (w.r.t.\ $P$) if there is no rule $R\in
\textit{ground}(P)$ for which the following three conditions all hold: (1)
$\textit{head}(R)\cap I\subseteq U$; (2) $\textit{body}^{+}(R)\subseteq
I\backslash U$; and (3) $\textit{body}^{-}(R)\cap I = \emptyset$. The
\emph{answer sets} of a program $P$ are the models of $P$ with no non-empty
unfounded subsets w.r.t.\ $P$.

Unless otherwise stated, in this paper the term ASP program is used to mean a
program consisting of a finite set of disjunctive rules.\footnote{The ILASP
systems support a wider range of ASP programs, including choice rules and
conditional literals, but we omit these concepts for simplicity.}

\subsection{Learning from Answer Sets}

The \emph{Learning from Answer Sets} framework, introduced
in~\cite{JELIAILASP}, is targeted at learning ASP programs. The basic framework
has been extended several times, allowing learning weak
constraints~\cite{ICLP15}, learning from context-dependent
examples~\cite{ICLP16} and learning from noisy examples~\cite{ACS18}. This
section presents the $\lasne$ learning framework, which is used in this
paper.\footnote{For details of how this approach can be extended to the full
$\loasne$ task, supported by ILASP, which enables the learning of weak
constraints, please see Appendix A.}

Examples in $\lasne$ are \emph{Context-dependent Partial Interpretations
(CDPIs)}. CDPIs specify what should or should not be an answer set of the
learned program. A \emph{partial interpretation} $e_{\mathit{pi}}$ is a pair of
sets of atoms $\langle e^{\mathit{inc}}, e^{\mathit{exc}}\rangle$ called the
inclusions and the exclusions, respectively. An interpretation $I$ extends
$e_{\mathit{pi}}$ if and only if $e^{\mathit{inc}}\subseteq I$ and
$e^{\mathit{exc}}\cap I = \emptyset$.\footnote{Note that partial
interpretations are very different to the examples given in many other ILP
approaches, which are usually atoms. A single positive example in $\lasne$ can
represent a full set of examples in a traditional ILP task (inclusions
correspond to traditional positive examples and exclusions correspond to
traditional negative examples). Multiple positive examples in $\lasne$ can be
used to learn programs with multiple answer sets, as each positive example can
be covered by a different answer set of the learned program. Negative examples
in $\lasne$ are used to express what should not be an answer set of the learned
program. For an in depth comparison of $\lasne$ with ASP-based ILP approaches
that use atomic examples, please see~\cite{AIJ17}.}
A \emph{Context-dependent Partial Interpretation} $e$ is a pair $\langle
e_{\mathit{pi}}, e_{\mathit{ctx}}\rangle$ where $e_{\mathit{pi}}$ is a partial
interpretation and $e_{\mathit{ctx}}$ (the \emph{context} of $e$) is a
disjunctive logic program. A program $P$ is said to \emph{accept} $e$ if there
is at least one answer set $A$ of $P\cup e_{\mathit{ctx}}$ that extends
$e_{\mathit{pi}}$ -- such an $A$ is called an \emph{accepting answer set} of
$e$ w.r.t.\ $P$, written $A\in \textit{AAS}(e, P)$.

\begin{example}

  Consider the program $P$, with the following two rules:

\begin{verbatim}
heads(V1) :- coin(V1), not tails(V1).
tails(V1) :- coin(V1), not heads(V1).
\end{verbatim}

  \begin{itemize}
    \item $P$ accepts $e = \langle \langle\lbrace \asp{heads(c1)}\rbrace,
      \lbrace\asp{tails(c1)}\rbrace\rangle, \lbrace
      \asp{coin(c1)\ruleend}\rbrace\rangle$. The only accepting answer set of
      $e$ \wrt $P$ is $\lbrace\asp{heads(c1)}, \asp{coin(c1)}\rbrace$.
    \item $P$ accepts $e = \langle \langle\lbrace \asp{heads(c1)}\rbrace$,
      $\lbrace\asp{tails(c1)}\rbrace\rangle, \lbrace
      \asp{coin(c1)\ruleend\;\;coin(c2)\ruleend}\rbrace\rangle$. The two
      accepting answer sets of $e$ \wrt $P$ are
      $\lbrace\asp{heads(c1)}$, $\asp{heads(c1)}$, $\asp{coin(c1)}$,
      $\asp{coin(c2)}\rbrace$ and $\lbrace\asp{heads(c1)}$, $\asp{tails(c1)}$,
      $\asp{coin(c1)}$, $\asp{coin(c2)}\rbrace$.
    \item $P$ does not accept $e = \langle \langle\lbrace \asp{heads(c1)},
      \asp{tails(c1)}\rbrace, \emptyset\rangle, \lbrace
      \asp{coin(c1)\ruleend}\rbrace\rangle$.
    \item $P$ does not accept $e = \langle \langle\lbrace \asp{heads(c1)}
      \rbrace, \lbrace\asp{tails(c1)}\rbrace\rangle, \emptyset\rangle$.
  \end{itemize}

\end{example}

In learning from answer sets tasks, CDPIs are given as either \emph{positive}
(resp.\ \emph{negative}) examples, which should (resp.\ should not) be accepted
by the learned program.

\paragraph{Noisy examples.}
In settings where all examples are correctly labelled (i.e.\ there is no
noise), ILP systems search for a hypothesis that covers all of the examples.
Many systems search for the optimal such hypothesis -- this is usually defined
as the hypothesis minimising the number of literals in $H$ ($|H|$). In real
settings, examples are often not guaranteed to be correctly labelled. In these
cases, many ILP systems (including ILASP) assign a penalty to each example,
which is the \emph{cost} of not covering the example. A CDPI $e$ can be
upgraded to a \emph{weighted CDPI} by adding a penalty $e_{\mathit{pen}}$,
which is either a positive integer or $\infty$, and a unique identifier,
$e_{\mathit{id}}$, for the example.

\paragraph{Learning Task.} Definition~\ref{def:lnas} formalises the $\lasne$
\emph{learning task}, which is the input of the ILASP systems. A
\emph{rule space} $S_{M}$ (often called a hypothesis space and characterised by
a \emph{mode bias}\footnote{We omit details of mode biases, as they are not
necessary to understand the rest of this paper. For details of the mode biases
supported in ILASP, please see the ILASP manual at
\url{https://doc.ilasp.com/}.} $M$) is a finite set of disjunctive rules,
defining the space of programs that are allowed to be learned. Given a rule
space $S_{M}$, a \emph{hypothesis} $H$ is any subset of $S_{M}$. The goal of a
system for $\lasne$ is to find an \emph{optimal inductive solution} $H$, which
is a subset of a given rule space $S_M$, that covers every example with
infinite penalty and minimises the sum of the length of $H$ plus the penalty of
each uncovered example.

\begin{definition}\label{def:lnas}
  An $\lasne$ task $T$ is a tuple of the form $\langle B, S_{M}, \langle E^{+},
  E^{-}\rangle\rangle$, where $B$ is an ASP program called the \emph{background
  knowledge}, $S_{M}$ is a rule space and $E^{+}$ and $E^{-}$ are (finite) sets
  of weighted CDPIs. Given a hypothesis $H \subseteq S_{M}$,

  \begin{enumerate}
    \item $\mathcal{U}(H, T)$ is the set consisting of: (a) all positive
      examples $e \in E^{+}$ such that $B \cup H$ does not accept $e$; and (b)
      all negative examples $e \in E^{-}$ such that $B \cup H$ accepts $e$.
    \item
      the score of $H$, denoted as $\mathcal{S}(H, T)$, is the sum $|H| +
      \sum_{e \in \mathcal{U}(H, T)} e_{pen}$.
    \item $H$ is an inductive solution of $T$ (written $H \in \lasne(T)$) if
      and only if $\mathcal{S}(H, T)$ is finite (i.e.\ $H$ must cover all
      examples with infinite penalty).
    \item $H$ is an \emph{optimal inductive solution} of $T$ (written $H \in$
      $^*\lasne(T)$) if and only if $\mathcal{S}(H, T)$ is finite and
      $\nexists H' \subseteq S_{M}$ such that $\mathcal{S}(H, T) >
      \mathcal{S}(H', T)$.
  \end{enumerate}
\end{definition}

\section{Conflict-driven Inductive Logic Programming}
\label{sec:cdilp}

In this section, we present ILASP's \emph{Conflict-driven ILP} (CDILP)
algorithm. For simplicity, we first explain how the CDILP approach solves
non-noisy learning tasks. In this case, the CDILP process iteratively builds a
set of \emph{coverage constraints} for each example, which specify certain
conditions for that example to be covered. In each iteration, the CDILP process
computes an optimal hypothesis $H^*$ (i.e. a hypothesis which is as short as
possible) that conforms to the existing coverage constraints; it then searches
for an example that is not covered by $H^*$ and computes a new coverage
constraint for that example -- esentially, this can be viewed as an explanation
of why $H^*$ is not an inductive solution. Eventually, the CDILP process
reaches an iteration where the hypothesis $H^*$ covers every example. In this
(final) iteration, $H^*$ is returned as the optimal solution of the task.
The noisy case is slightly more complicated. Firstly, the computed optimal
hypothesis $H^*$ does not need to cover every example and, therefore, does not
need to conform to every coverage constraint. Instead, $H^*$ must be optimal in
terms of its length plus the penalties of all examples for which $H^*$ does not
cover at least one existing coverage constraint -- essentially, this search
``chooses'' not to cover certain examples. Secondly, the search for an
uncovered example must find an example that the hypothesis search did not
``choose'' not to cover (i.e. an uncovered example $e$ such that $H^*$ conforms
to every existing coverage constraint for $e$).

Roughly speaking, coverage constraints are boolean constraints over the rules
that a hypothesis must contain to cover a particular example; for example, they
may specify that a hypothesis must contain at least one of a particular set of
rules and none of another set of rules.
The coverage constraints supported by ILASP are formalised by
Definition~\ref{def:covf}.
Throughout the rest of the paper, we assume $T=\langle B, S_M, E\rangle$ to be
an $ILP_{LAS}^{noise}$ learning task. We also assume that every rule $R$ in
$S_M$ has a unique identifier, written $R_{\mathit{id}}$.

\begin{definition}\label{def:covf}
  Let $S_M$ be a rule space. A \emph{coverage formula} over $S_M$ takes one of
  the following forms:
  \begin{itemize}
    \item $R_{id}$, for some $R\in S_M$.
    \item $\lnot F$, where $F$ is a coverage formula over $S_M$.
    \item $F_1\lor\ldots\lor F_n$, where $F_1,\ldots,F_n$ are coverage formulas over $S_M$.
    \item $F_1\land\ldots\land F_n$, where $F_1,\ldots,F_n$ are coverage formulas over $S_M$.
  \end{itemize}

  The semantics of coverage formulas are defined as follows. Given a hypothesis $H$:

  \begin{itemize}
    \item $R_{id}$ accepts $H$ if and only if $R \in H$.
    \item $\lnot F$ accepts $H$ if and only if $F$ does not accept $H$.
    \item $F_1\lor\ldots\lor F_n$ accepts $H$ if and only if $\exists i\in [1,n]$ s.t. $F_i$ accepts $H$.
    \item $F_1\land\ldots\land F_n$ accepts $H$ if and only if $\forall i\in [1,n]$ s.t. $F_i$ accepts $H$.
  \end{itemize}

  A \emph{coverage constraint} is a pair $\langle e, F\rangle$, where $e$ is an
  example in $E$ and $F$ is a coverage formula, such that for any $H\subseteq
  S_M$, if $e$ is covered then $F$ accepts $H$.

\end{definition}

\begin{example}\label{eg:covf}
  Consider a task with background knowledge $B$ and the following rule
  space $S_M$:\\[-2mm]

  \noindent
  $\left\{\begin{array}{llllll}
    h^1:&\mbox{\hspace{-2mm}}\asp{heads\ruleend}                   &\mbox{\hspace{-2mm}}
    h^2:&\mbox{\hspace{-2mm}}\asp{tails\ruleend}                   &\mbox{\hspace{-2mm}}
    h^3:&\mbox{\hspace{-2mm}}\asp{heads\codeif tails\ruleend}      \\
    h^4:&\mbox{\hspace{-2mm}}\asp{tails\codeif heads\ruleend}      &\mbox{\hspace{-2mm}}
    h^5:&\mbox{\hspace{-2mm}}\asp{heads\codeif \naf tails\ruleend} &\mbox{\hspace{-2mm}}
    h^6:&\mbox{\hspace{-2mm}}\asp{tails\codeif \naf heads\ruleend} \\
  \end{array}\right\}$\\[-1mm]

  Let $e$ be the positive CDPI example $\langle\langle \lbrace
  \asp{heads}\rbrace, \lbrace \asp{tails}\rbrace\rangle, \emptyset\rangle$. For
  any hypothesis $H\subseteq S_M$, $H$ can only cover $e$ if $H$ contains at
  least one rule defining $\asp{head}$. So $H$ must contain either $h^1$, $h^3$
  or $h^5$. This is captured by the coverage constraint $\langle e,
  h^1_{id}\lor h^3_{id} \lor h^5_{id}\rangle$.

  Note that not every hypothesis that is accepted by the coverage constraint
  covers $e$. For instance the hypothesis $\lbrace h^1, h^2\rbrace$ does not
  (its only answer set contains $\asp{tails}$). However, every hypothesis that
  does cover $e$ conforms to the coverage constraint (which exactly the
  condition given in Definition~\ref{def:covf}).

  Coverage constraints are not necessarily unique. There are usually many
  coverage constraints that could be computed for each example, and in fact the
  method for deriving a coverage constraint is a modular part of the CDILP
  procedure (detailed in the next sub-section). An alternative coverage
  constraint that could be computed in this case is $\langle e, (h^1_{id} \lor
  h^5_{id})\land \lnot h^2 \land \lnot h^4 \rangle$.
\end{example}

In each iteration of the CDILP procedure, a set of coverage constraints
$\textit{CC}$ is \emph{solved}, yielding: (1) a hypothesis $H$ which is optimal
\wrt $\textit{CC}$ (i.e. the length of $H$ plus the penalties of examples $e$
for which there is at least one coverage constraint $\langle e, F\rangle$ s.t.\
$F$ does not accept $H$); (2) a set of examples $U$ which are known not to be
covered by $H$; and (3) a score $s$ which gives the score of $H$, according to
the coverage constraints in $\textit{CC}$. These three elements, $H$, $U$ and
$s$ form a \emph{solve result}, which is formalised by the following
definition.

\begin{definition}
  Let $CC$ be a set of coverage constraints. A \emph{solve result} is a tuple
  $\langle H, U, s\rangle$, such that:

  \begin{enumerate}
    \item $H \subseteq S_M$;
    \item $U$ is the set of examples $e$ (of any type) in $E$ for which there
      is at least one coverage constraint $\langle e, F\rangle$ such that $F$
      does not accept $H$;
    \item $s = |H| + \sum\limits_{u\in U} u_{pen}$;
    \item $s$ is finite.
  \end{enumerate}

  A solve result $\langle H, U, s\rangle$ is said to be \emph{optimal} if there
  is no solve result $\langle H', U', s'\rangle$ such that $s > s'$.
\end{definition}

Theorem~\ref{thm:lt} shows that for any solve result $\langle H, U, s\rangle$,
every example in $U$ is not covered by $H$ and $s$ is a lowerbound for the
score of $H$. Also, $s$ is equal to the score of $H$ if and only if $U$ is
exactly the set of examples that are not covered by $H$.

\appendixTheorem{lt}{
  Let $CC$ be a set of coverage constraints. For any solve result $\langle H,
  U, s\rangle$, $U \subseteq \mathcal{U}(H, T)$ and $s \leq \mathcal{S}(H, T)$.
  Furthermore, $s = \mathcal{S}(H, T)$ if and only if $U = \mathcal{U}(H, T)$.
}{

  Let $u$ be an arbitrary example in $U$. There must be at least one coverage
  constraint $\langle u, F\rangle \in CC$ such that $H$ is not accepted by $F$.
  By definition of $\langle u, F\rangle$ being a coverage constraint, any
  hypothesis that is not accepted by $F$ cannot cover $u$; hence, $H$ does not
  cover $u$, and thus $u \in \mathcal{U}(H, T)$. Therefore, $U \subseteq
  \mathcal{U}(H, T)$.
  This implies that $\sum\limits_{u \in U} u_{pen} \leq \sum\limits_{u \in
  \mathcal{U}(H, T)} u_{pen}$, and hence, $s \leq \mathcal{S}(H, T)$.

  It remains to show that $s = \mathcal{S}(H, T)$ if and only if $U =
  \mathcal{U}(H, T)$.  We can show this by demonstrating that $U\neq
  \mathcal{U}(H, T)$ if and only if $s\neq \mathcal{S}(H, T)$. Assume that
  $U\neq \mathcal{U}(H, T)$. This holds if and only if $U\subset \mathcal{U}(H,
  T)$, and hence, if and only if $s < \mathcal{S}(H,
  T)$. This holds if and only if $s \neq \mathcal{S}(H, T)$.
}

A crucial consequence of Theorem~\ref{thm:lt} (formalised by
Corollary~\ref{corr:re}) is that if $H$ is not an optimal inductive solution,
then for any optimal solve result containing $H$ there will be at least one
counterexample to $H$ (i.e.\ an example that is not covered by $H$) that is not
in $U$. This means that when a solve result is found such that $U$ contains
every example that is not covered by $H$, $H$ is guaranteed to be an optimal
inductive solution of $T$. This is used as the termination condition for the
CDILP procedure in the next section.

\begin{corollary}\label{corr:re}
  Let $CC$ be a set of coverage constraints. For any optimal solve result
  $\langle H, U, s\rangle$, such that $H$ is not an optimal solution of $T$,
  there is at least one counterexample to $H$ that is not in $U$.
\end{corollary}

\subsection{The CDILP Procedure}

\newcommand{\relfinder}[0]{
  \begin{tikzpicture}
    \node (0) at (0, 0) {Counterexample Search};
    \node [draw,thick,minimum width=3cm] (1) at (0, -1) {Inputs: $\langle H, U, s\rangle$, $T$};
    \node [draw,thick,minimum width=1.5cm] (6) at (-2, -2.5) {\begin{tikzpicture}
      \node [minimum width=3cm] (3) at (0, 0) {Return counterexample};
      \node [minimum width=3cm] (4) at (0, -0.5) {$\textit{ce}$ if one exists};
    \end{tikzpicture}};
    \node [draw,thick,minimum width=1.5cm] (8) at (2, -2.5) {\begin{tikzpicture}
      \node [minimum width=3cm] (3) at (0, 0) {Otherwise, return};
      \node [minimum width=3cm] (4) at (0, -0.5) {$H$ as learned program};
    \end{tikzpicture}};
    \node [minimum width=3cm] (7) at (0, -3.3) {};
    \draw [->] (1) to (6) {};
    \draw [->] (1) to (8) {};
  \end{tikzpicture}
}

\newcommand{\hypsearch}[0]{
  \begin{tikzpicture}
    \node (0) at (0, 0) {Hypothesis Search};
    \node [draw,thick,minimum width=3cm] (1) at (0, -1) {Inputs: Coverage Constraints $\textit{CC}$};
    \node [draw,thick,minimum width=1.5cm] (6) at (-1.75, -2.5) {\begin{tikzpicture}
      \node [minimum width=3cm] (3) at (0, 0) {If $\textit{CC}$ is satisfiable, update};
      \node [minimum width=3cm] (4) at (0, -0.5) {optimal solve result $\langle H, U, s\rangle$};
    \end{tikzpicture}};
    \node [draw,thick,minimum width=1.5cm] (8) at (2.75, -2.5) {\begin{tikzpicture}
      \node [minimum width=3cm] (3) at (0, 0) {Otherwise, return};
      \node [minimum width=3cm] (4) at (0, -0.5) {\texttt{UNSATISFIABLE}};
    \end{tikzpicture}};
    \node [minimum width=3cm] (7) at (0, -3.3) {};
    \draw [->] (1) to (6) {};
    \draw [->] (1) to (8) {};
  \end{tikzpicture}
}

\newcommand{\translator}[0]{
  \begin{tikzpicture}
    \node (0) at (0, 0) {Conflict Analysis};
    \node [draw,thick,minimum width=3cm] (1) at (0, -1) {Inputs: $\textit{ce}, H, T$};
    \node [draw,thick,minimum width=1.5cm] (2) at (0, -2.5) {\begin{tikzpicture}
      \node [minimum width=3cm] (3) at (0, 0) {Compute $F$ s.t.\ $\langle \textit{ce}, F\rangle$ is a coverage constraint.};
      \node [minimum width=3cm] (4) at (0, -0.5) {Add $\langle \textit{ce}, F\rangle$ to $\textit{CC}$};
    \end{tikzpicture}};
    \node [minimum width=3cm] (7) at (0, -3.3) {};
    \draw [->] (1) to (2) {};
  \end{tikzpicture}
}

\newcommand{\implication}[0]{
  \begin{tikzpicture}
    \node (0) at (0, 0) {Constraint Propagation};
    \node [draw,thick,minimum width=3cm] (1) at (0, -1) {Inputs: $F$, $T$};
    \node [draw,thick,minimum width=1.5cm] (2) at (0, -2.5) {\begin{tikzpicture}
      \node [minimum width=3cm] (3) at (0, 0) {Compute examples $e$ s.t.\ $\langle e, F\rangle$ is a coverage constraint.};
      \node [minimum width=3cm] (4) at (0, -0.5) {For each computed $e$, add $\langle e, F\rangle$ to $\textit{CC}$};
    \end{tikzpicture}};
    \node [minimum width=3cm] (7) at (0, -3.3) {};
    \draw [->] (1) to (2) {};
  \end{tikzpicture}
}

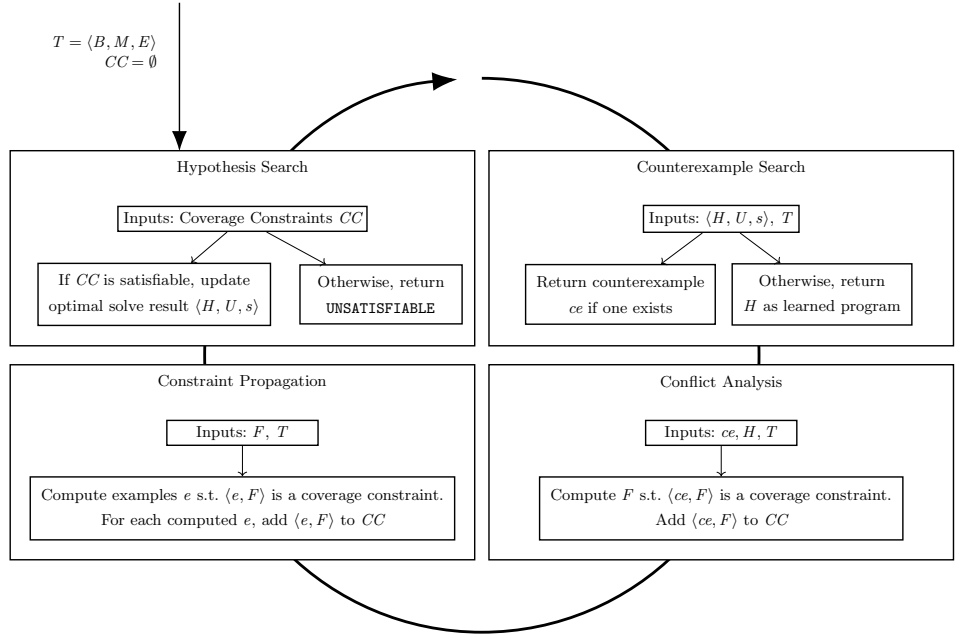
\begin{figure}[t]
  \begin{center}
    \scalebox{0.67}{
    \begin{tikzpicture}
      \draw [ultra thick,-{Latex[length=5mm]}] (0, 1) arc(90:-265:5.5);
      \draw [thick,-{Latex[length=4mm]}] (-6, 2.5) to (-6, -0.45) {};
      \node [draw,thick,minimum width=9cm,fill=white] (c) at (-4.75, -2.375) {\hypsearch};
      \node [draw,thick,minimum width=9cm,fill=white] (d) at (4.75, -2.375) {\relfinder};
      \node [draw,thick,minimum width=9cm,fill=white] (t) at (-4.75, -6.625) {\implication};
      \node [draw,thick,minimum width=9cm,fill=white] (v) at (4.75, -6.625) {\translator};
      \node (i) at (-7.5, 1.5) {\begin{tikzpicture}
        \node (k) at (0, 0) {
          $\begin{array}{r}
          T=\langle B, M, E\rangle\\
          \textit{CC}=\emptyset\\
        \end{array}$};
      \end{tikzpicture}};
    \end{tikzpicture}
    }
  \end{center}
  \caption{\label{fig:cdilp} The ILASP CDILP procedure.}
\end{figure}

The algorithm presented in this section is a cycle comprised of four steps,
illustrated in Figure~\ref{fig:cdilp}. Step 1, the \emph{hypothesis search},
computes an optimal solve result $\langle H, U, s\rangle$ w.r.t.\ the current
set of coverage constraints. Step 2, the \emph{counterexample search}, finds an
example $e$ which is not in $U$ (i.e.\ an example whose coverage constraints
are respected by $H$) that $H$ does not cover. The existence of such an example
$e$ is called a \emph{conflict}, and shows that the coverage constraints are
incomplete.  The third step, \emph{conflict analysis}, resolves the situation
by computing a new coverage constraint for $e$ that is not respected by $H$.
The fourth step, \emph{constraint propagation}, is optional and only useful for
noisy tasks. The idea is to check whether the newly computed coverage
constraint can also be used for other examples, thus ``boosting'' the penalty
that must be paid by any hypothesis that does not respect the coverage
constraint and reducing the number of iterations of the CDILP procedure. The
CDILP procedure is formalised by Algorithm~\ref{alg:CDILP}.

\algnewcommand\algorithmicforeach{\textbf{for each}}
\algdef{S}[FOR]{ForEach}[1]{\algorithmicforeach\ #1\ \algorithmicdo}

\begin{algorithm}
  \begin{algorithmic}[1]
    \Procedure{CDILP}{$T$}
      \State{$CC = \emptyset;$}
      \State{$solve\_result = hypothesis\_search(CC);$}
      \While{$solve\_result \neq \asp{nil}$}
        \State{$\langle H, U, s\rangle = solve\_result;$}
        \State{$ce = counterexample\_search(solve\_result, T);$}
        \If{$ce == \asp{nil}$}
          \State{\textbf{return}\,$H;$}
        \Else
          \State{$F = conflict\_analysis(ce, H, T);$}
          \State{$CC\ruleend insert(\langle ce, F\rangle)$}
          \State{$prop\_egs = propagate\_constraints(F, T);$}
          \ForEach {$e \in prop\_egs$}
            \State{$CC\ruleend insert(\langle e, F\rangle)$}
          \EndFor
          \State{$solve\_result = hypothesis\_search(CC);$}
        \EndIf
      \EndWhile
      \State{\textbf{return}\,\,$\asp{UNSATISFIABLE};$}
    \EndProcedure
  \end{algorithmic}
  \caption{$CDILP(T)$ \label{alg:CDILP}}
\end{algorithm}

\paragraph{Hypothesis Search.} The hypothesis search phase of the CDILP
procedure finds an optimal solve result of the current $CC$ if one exists; if
none exists, it returns \texttt{nil}. This search is performed using Clingo. By
default, this process uses Clingo 5's~\cite{clingo5} C++ API to enable
multi-shot solving (adding any new coverage constraints to the program and
instructing the solver to continue from where it left off in the previous
iteration). This multi-shot solving can be disabled by calling ILASP with the
``\texttt{--restarts}'' flag.
This ASP program is entirely based on the coverage constraints and does not use
the examples or background knowledge. This means that if checking the coverage
of an example in the original object-level domain is computationally intensive
(e.g.\ if it has a large grounding or the decision problem is NP-hard or
higher), the hypothesis search phase can be much easier than solving a
meta-level encoding of the background knowledge and examples because the hard
aspects of the original problem have essentially been ``compiled away'' in the
computation of the coverage constraints.
For reasons of brevity, the actual ASP encodings
are omitted from this paper, but detailed descriptions of very similar ASP
encodings can be found in~\cite{ILASP_thesis}.

\paragraph{Counterexample Search.} A counterexample to a solve result $\langle
H, U, s\rangle$, is an example $e$ that is not covered by $H$ and is not in
$U$. The existence of such an example proves that the score $s$ is lower than
$\mathcal{S}(H, T)$, and hence, $H$ may not be an optimal solution of $T$. This
search is again performed using Clingo, and is identical to the
$findRelevantExample$ method of ILASP2i~\cite{ICLP16} and
ILASP3~\cite{ILASP_thesis}. If no counterexample exists, then by
Corollary~\ref{corr:re}, $H$ must be an optimal solution of $T$, and is
returned as such; if not, the procedure continues to the conflict analysis
step.

\paragraph{Conflict Analysis.} Let $\langle H, U, s\rangle$ be the most recent
solve result and $ce$ be the most recent counterexample. The goal of this step
is to compute a coverage formula $F$ that does not accept $H$ but that must
hold for $ce$ to be covered. The coverage constraint $\langle ce, F\rangle$ is
then added to $CC$. This means that if $H$ is computed in a solve result in the
hypothesis search phase of a future iteration, then it will be guaranteed to be
found with a higher score (including the penalty paid for not covering $ce$).
There are many possible strategies for performing conflict analysis, several of
which are presented in the next section and evaluated in
Section~\ref{sec:eval}. Beginning with ILASP version 4.0.0, the ILASP system
allows a user to customise the learning process by providing a Python script
(called a \emph{PyLASP} script). Future versions of ILASP will likely
contain many strategies, appropriate for different domains and different kinds
of learning task. In particular, this allows a user to define their own
conflict analysis methods. Provided the conflict analysis method is guaranteed
to terminate and compute a coverage constraint whose coverage formula does not
accept the most recent hypothesis, the customised CDILP procedure is guaranteed
to terminate and return an optimal solution of $T$ (resources permitting). We
call such a conflict analysis method \emph{valid}. The three conflict analysis
methods presented in the next section are proven to be valid.

\paragraph{Constraint Propagation.} The final step is optional (and can be
disabled in ILASP with the flag ``\texttt{-ncp}''). In a task with many
examples with low penalties, but for which the optimal solution has a high
score, there are likely to be many iterations required before the hypothesis
search phase finds an optimal solution. This is because each new coverage
constraint only indicates that the next hypothesis computed should either
conform to the coverage constraint, or pay a very small penalty. The goal of
constraint propagation is to find a set of examples which are guaranteed to not
be covered by any hypothesis $H$ that is not accepted by $F$. For each such
example $e$, $\langle e, F\rangle$ can be added as a coverage constraint (this
is called \emph{propagating} the constraint to $e$). Any solve result
containing a hypothesis that does not conform to $F$ must pay the penalty not
only for the counterexample $ce$, but also for every constraint that $F$ was
propagated to. In Section~\ref{sec:eval}, it is shown that by lowering the
number of iterations required to solve a task, constraint propagation can
greatly reduce the overall execution time.

There are two methods of constraint propagation supported in the current
version of ILASP. Both were used in ILASP3 and described in detail
in~\cite{ILASP_thesis} as ``implication'' and ``propagation'', respectively.
The first is used for positive examples and brave ordering examples, and for
each example $e$ searches for a hypothesis that is not accepted by $F$, but
that covers $e$. If none exists, then the constraint can be propagated to $e$.
The second method, for negative examples, searches for an accepting answer set
of $e$ that is guaranteed to be an answer set of $B\cup H\cup e_{ctx}$ for any
hypothesis $H$ that is accepted by $F$. A similar method is possible for
propagating constraints to cautious orderings; however, our initial experiments
have shown it to be ineffective in practice, as although it does bring down the
number of iterations required, it adds more computation time than it saves.
Similarly to the conflict analysis phase, users can provide their own strategy
for constraint propagation in PyLASP, and in future versions of ILASP will
likely have a range of alternative constraint propagation strategies built in.

\paragraph{Correctness of CDILP.}

Theorem~\ref{thm:sac} proves the correctness of the CDILP approach. The proof
of the theorem assumes that the conflict analysis method is valid (which is
proven for the three conflict analsysis methods presented in the next section).
A \emph{well-formed} task has finite number of examples, and for each example
context $C$, the program $B\cup S_{M}\cup C$ has a finite grounding. The
theorem shows that the CDILP approach is guaranteed to terminate, and is both
sound and complete \wrt the optimal solutions of a task; i.e.\ any hypothesis
returned is guaranteed to be an optimal solution and if at least one solution
exists, then CDILP is guaranteed to return an optimal solution.

\appendixTheorem{sac}{
  For any $ILP_{LAS}^{noise}$ well-formed task $T$, $CDILP(T)$ is guaranteed
  to terminate and return an optimal solution of $T$ if $T$ is satisfiable, and
  return \texttt{UNSATISFIABLE} otherwise.
}{
  As there is a finite number of hypotheses $H\subseteq S_M$ and a finite
  number of sets of examples $U$ from the task $T$, there is a finite number
  of solve results.  Hence, to demonstrate that $CDILP(T)$ terminates, it
  suffices to show that the hypothesis search phase cannot produce the same
  solve result in two iterations (demonstrating that there must be a finite
  number of iterations).  Consider an arbitrary iteration with a solve result
  $\langle H, U, s\rangle$. As the conflict analysis method is valid, it must
  produce a coverage formula $F$ such that $H$ is not accepted by $F$. As
  $\langle u, F\rangle$ is added as a coverage constraint, for any future solve
  result $\langle H', U', s'\rangle$ such that $H'=H$, $U'$ must contain $u$.
  Thus, as $u\not\in U$, the solve result $\langle H, U, s\rangle$ cannot be
  produced by the hypothesis search phase of any future iteration. Hence,
  $CDILP(T)$ must terminate.

  Assume that $T$ is satisfiable. Then there must be at least one optimal
  solution $H^*$ of $T$. For any set of coverage constraints $CC$, $\langle
  H^*, U_{CC}, s_{CC}\rangle$ is a solve result of $CC$ (where $U_{CC}$ is the set
  of examples $u$ for which there is at least one $\langle u, F\rangle$ such
  that $F$ does not accept $H$ and $s_{CC}$ is the corresponding score).
  Hence, $CDILP(T)$ cannot return \texttt{UNSATISFIABLE}. As $CDILP(T)$
  terminates, this means that it must return a hypothesis $H$. Assume for
  contradiction that $CDILP(T)$ returns a suboptimal solution $H'$. In the
  final iteration, the solve result $\langle H', U', s'\rangle$ must be such
  that $U' = \mathcal{U}(H, T)$, and hence (by Theorem~\ref{thm:lt}), $s' =
  \mathcal{S}(H', T)$. As $H'$ is suboptimal, $\mathcal{S}(H', T) >
  \mathcal{S}(H^*, T)$. Hence $\langle H', U', s'\rangle$ could not have been
  an optimal solve result for $CC$, as $s_{CC}$ must be lower than $s'$. This
  contradiction proves that $CDILP(T)$ must return an optimal solution of $T$.

  It remains to show that if $T$ is unsatisfiable, $CDILP(T)$ returns
  \texttt{UNSATISFIABLE}. As $CDILP(T)$ terminates, it suffices to show that
  $CDILP(T)$ cannot return a hypothesis. Assume for contradiction that it does.
  Then there must be a set of coverage constraints $CC$ and a solve result
  $\langle H, U, s\rangle$ such that $U = \mathcal{U}(H, T)$. As $s$ is finite,
  by Theorem~\ref{thm:lt}, this shows that $H$ has a finite score. Hence, $H$
  is a solution of $T$, which contradicts that $T$ is unsatisfiable. Hence, if
  $T$ is unsatisfiable, $CDILP(T)$ must return \texttt{UNSATISFIABLE}.
}

\subsection{Comparison to previous ILASP systems}

ILASP1 and ILASP2 both encode the search for an inductive solution as a
meta-level ASP program. They are both iterative algorithms and use multi-shot
solving~\cite{clingo5} to add further definitions and constraints to the
meta-level program throughout the execution. However, the number of rules in
the grounding of the initial program is roughly proportional to the number of
rules in the grounding of $B\cup S_M$ (together with the rules in the contexts
of each example) multiplied by the number of positive examples and (twice the
number of) brave orderings~\cite{ILASP_thesis}. This means that neither ILASP1
nor ILASP2 scales well w.r.t.\ the number of examples.

ILASP2i attempts to remedy the scalability issues of ILASP2 by iteratively
constructing a set of \emph{relevant} examples. The procedure is similar to
CDILP in that it searches (using ILASP2) for a hypothesis that covers the
current set of relevant examples, and then searches for a counterexample to the
current hypothesis, which is then added to the set of relevant examples before
the next hypothesis search. This simple approach allows ILASP2i to scale to
tasks with large numbers of examples, providing the final set of relevant
examples stays relatively small~\cite{ICLP16}. However, as ILASP2i uses ILASP2
for the hypothesis search, it is still using a meta-level ASP program which has
a grounding that is proportional to the number of relevant examples, meaning
that if the number of relevant examples is large, the scalability issues
remain.

The CDILP approach defined in this section goes further than ILASP2i in that
the hypothesis search phase is now completely separate from the groundings of
the rules in the original task. Instead, the program used by the hypothesis
search phase only needs to represent the set of (propositional) coverage
formulas.

The major advantage of the CDILP approach compared to ILASP2i (as demonstrated
by the evaluation in Section 5) is on tasks with noisy examples. Tasks with
noisy examples are likely to lead to a large number of relevant examples. This
is because if a relevant example has a penalty it does not need to be covered
during the hypothesis search phase. It may be required that a large number of
``similar'' examples are added to the relevant example set before any of the
examples are covered. The CDILP approach overcomes this using constraint
propagation. When the first example is found, it will be propagated to all
``similar'' examples which are not covered for the same ``reason''. In the next
iteration the hypothesis search phase will either have to attempt to cover the
example, or pay the penalty of all of the similar examples.

\section{Conflict Analysis}
\label{sec:conflict}
This section presents the three approaches to conflict analysis available in
the ILASP system. Each approach relies on the notion of a \emph{translation}
of an example, which is formalised in the next sub-section. A translation is a
coverage formula accepts exactly those hypotheses which cover the example;
i.e.\ the coverage formula is both necessary and sufficient for the example to
be covered.

One approach to conflict analysis is to compute a full translation of an
example in full and return this coverage formula. This is, in fact, the method
used by the ILASP3 algorithm. However, as this operation can be extremely
expensive, and can lead to an extremely large coverage formula it may not be
the best approach. The other two conflict analysis techniques (available in
ILASP4) compute shorter (and less specific) coverage formulas, which are only
necessary for the example to be covered. This may mean that the same
counterexample is found in multiple iterations of the CDILP procedure (which
cannot occur in ILASP3), but for each iteration, the conflict analysis phase is
usually significantly cheaper. The evaluation in Section~\ref{sec:eval}
demonstrates that although the number of iterations in ILASP4 is likely to be
higher than in ILASP3, the overall running time is often much lower.


\subsection{Positive CDPI examples}

In this section, we describe how the three conflict analysis methods behave
when $\textit{conflict\_analysis}(e, H, T)$ is called for a positive CDPI
example $e$, a hypothesis $H$ (which does not accept $e$) and a learning task
$T$.
Each of the three conflict analysis methods presented in this paper work by
incrementally building a coverage formula which is a disjunction of the form
$D_1\lor\ldots\lor D_n$. The intuition is that in the $i^{th}$ iteration the
algorithm searches for a hypothesis $H'$ that accepts $e$, but which is not
accepted by $D_1\lor\ldots\lor D_{i-1}$. The algorithm then computes an answer
set $I$ of $B\cup e_{ctx}\cup H'$ that extends $e_{pi}$.
If such an $H'$ exists, a coverage formula $D_i$ is computed (and added to the
disjunction) s.t.\ $D_i$ does not accept $H$ and $D_i$ is necessary for $I$ to
be an accepting answer set of $e$ (i.e.\ $D_i$ is respected by every hypothesis
that accepts $e$). We denote this formula $\psi(I, e, H, T)$.
If no such $H'$ exists then $F = D_1\lor\ldots\lor D_{i-1}$ is necessary for
the example to be accepted and it will clearly not accept $H$ (as none of the
disjuncts accept $H$). It can therefore be returned as a result of the conflict
analysis.

\begin{algorithm}
  \begin{algorithmic}[1]
    \Procedure{iterative\_conflict\_analysis}{$e, H, T, \psi$}
      \State{$F = \bot$;}
      \While{$\exists I$, $\exists H' \subseteq S_M$ s.t.\ $F$ does not accept $H'$ and $I\in AAS(e, B\cup H')$}
        \State{Fix an arbitrary such $I$}
        \State{$F = F \lor \psi(I, e, H, T)$;}
      \EndWhile
      \State{\textbf{return}\,$F$;}
    \EndProcedure
  \end{algorithmic}
  \caption{$\textit{iterative\_conflict\_analysis}(e, H, T, \psi)$ \label{alg:ca_cdpi}}
\end{algorithm}

The overall conflict analysis methods are formalised by
Algorithm~\ref{alg:ca_cdpi}. The three conflict analysis methods use different
definitions of $\psi(I, e, H, T)$, which are given later this section.
Each version of $\psi(I, e, H, T)$ is linked to the notion of the
\emph{translation} of an interpretation $I$. Essentially, this is a
coverage formula that is accepted by exactly those hypotheses $H'$ for which
$I$ is an answer set of $AS(B\cup e_{ctx}\cup H')$. The translation is composed
of two parts: first, a the set of rules which must not appear in $H'$
for $I$ to be an answer set -- these are the rules for which $I$ is not a
model; and second, a set of disjunctions of rules -- for each disjunction, at
least one of the rules must appear in $H'$ for $I$ to be an answer set. The
notion of the translation of an interpretation is closely related to the
definition of an answer set based on unfounded sets. For $I$ to be an answer
set, it must be a model of $H$ and it must have no non-empty unfounded subsets
w.r.t.\ $B\cup e_{ctx}\cup H'$. For any $H'$ that is accepted by the
translation, the first part guarantees that $I$ is a model of $H'$, while the
second part guarantees that there are no non-empty unfounded subsets of $I$, by
ensuring that for each potential non-empty unfounded subset $U$ there is at
least one rule in $H'$ that prevents $U$ from being unfounded.

\begin{definition}
  Let $I$ be an interpretation and $e$ be a CDPI. The \emph{translation} of
  $\langle I, e\rangle$ (denoted $\mathcal{T}(I, e, T)$) is the coverage
  formula constructed by taking the conjunction of the following coverage
  formulas:

  \begin{enumerate}
    \item $\lnot R_{id}$ for each $R \in S_M$ such that $I$ is not a model of
      $R$.
    \item $R_{id}^1\lor\ldots\lor R_{id}^n$ for each subset minimal set of
      rules $\lbrace R^1,\ldots,R^n\rbrace$ such that there is at least one
      non-empty unfounded subset of $I$ \wrt $B\cup e_{ctx}\cup
      (S_{M}\backslash \lbrace R^1,\ldots,R^n\rbrace)$.
  \end{enumerate}

  \noindent 
  We write $\mathcal{T}_1(I, e, T)$ and $\mathcal{T}_2(I, e, T)$ to refer to
  the conjunctions of coverage formulas in (1) and (2), respectively. Note that
  the empty conjunction is equal to $\top$.
\end{definition}

\begin{example}\label{eg:translate}
  Consider an $\lasne$ task $T$ with background knowledge $B$ and hypothesis
  space $S_M$ as defined below.

  \begin{multicols}{2}

    $B = \left\{\begin{array}{l}
      \asp{p \codeif \naf q.}\\
      \asp{q \codeif \naf p.}\\
    \end{array}\right\}$

    $S_M = \left\{ \begin{array}{rl}
        h^1:& \asp{r \codeif t.}\\
        h^2:& \asp{t \codeif q.}\\
        h^3:& \asp{r.}\\
        h^4:& \asp{s \codeif t.}\\
      \end{array}
    \right\}$

  \end{multicols}

  Let $e$ be a CDPI such that $e_{pi} = \langle \lbrace \asp{r}\rbrace,
  \emptyset\rangle$, and $e_{ctx} = \emptyset$. Consider the four
  interpretations
  $I_1 = \lbrace \asp{q}$, $\asp{r}$, $\asp{t}\rbrace$,
  $I_2 = \lbrace \asp{q}$, $\asp{r}\rbrace$,
  $I_3 = \lbrace \asp{p}$, $\asp{r}\rbrace$ and
  $I_4 = \lbrace \asp{q}$, $\asp{r}$, $\asp{s}$, $\asp{t}\rbrace$.
  The translations are as follows:

  \begin{itemize}
    \item $\mathcal{T}(I_1, e, T) = \lnot h^4_{id} \land (h^1_{id} \lor h^3_{id}) \land h^2_{id}$.
    \item $\mathcal{T}(I_2, e, T) = \lnot h^2_{id} \land h^3_{id}$.
    \item $\mathcal{T}(I_3, e, T) = h^3_{id}$.
    \item $\mathcal{T}(I_4, e, T) = h^4_{id} \land (h^1_{id} \lor h^3_{id}) \land h^2_{id}$.
  \end{itemize}
\end{example}

The following theorem shows that for any interpretation $I$ that extends
$e_{pi}$, the translation of $I$ w.r.t.\ $e$ is a coverage formula that captures
the class of hypotheses $H$ for which $I$ is an accepting answer set of $B\cup
H$ w.r.t.\ $e$.

\appendixTheorem{translateCDPI}{
  Let $e$ be a CDPI and $I$ be a model of $B\cup e_{ctx}$ that extends
  $e_{pi}$. For any hypothesis $H\subseteq S_{M}$, $I \in AAS(e, B \cup H)$ if
  and only if the translation of $\langle I, e\rangle$ accepts $H$.
}{
  Assume $I \in AAS(e, B\cup H)$.

  \proofindent{
    $\Leftrightarrow$ $I \in AS(B\cup H \cup e_{ctx})$ (as $I$ is assumed to extend $e_{pi}$).

    $\Leftrightarrow I$ is a model of $B\cup H\cup e_{ctx}$ and there are no
    non-empty subsets of $I$ \wrt $B\cup H\cup e_{ctx}$

    $\Leftrightarrow I$ is a model of $H$ and there are no non-empty subsets of
    $I$ \wrt $B\cup H\cup e_{ctx}$ (as $I$ is a model of $B\cup e_{ctx}$)

    $\Leftrightarrow I$ is a model of $H$ and for each set of rules $\lbrace
    R^1,\ldots,R^n\rbrace$ s.t.\ there is a non-empty unfounded subset of $I$
    \wrt $B\cup e_{ctx}\cup (S_{M}\backslash \lbrace R^1,\ldots,R^n\rbrace)$,
    $H\cap \lbrace R^1,\ldots,R^n\rbrace \neq \emptyset$.

    $\Leftrightarrow I$ is a model of $H$ and for each subset minimal set of
    rules $\lbrace R^1,\ldots,R^n\rbrace$ s.t.\ there is a non-empty unfounded
    subset of $I$ \wrt $B\cup e_{ctx}\cup (S_{M}\backslash \lbrace
    R^1,\ldots,R^n\rbrace)$, $H\cap \lbrace R^1,\ldots,R^n\rbrace \neq
    \emptyset$.

    $\Leftrightarrow$ the translation of $\langle I, e\rangle$ accepts $H$.
  }
}

Given the notion of a translation, one potential option for defining $\psi(I,
e, H, T)$ would be to let $\psi(I, e, H, T) = \mathcal{T}(I, e, T)$. In
fact, this is exactly the approach adopted by the ILASP3 algorithm. We show
later in this section that this results in a valid method for conflict
analysis; however, the coverage constraints returned from this method can be
extremely long, and computing them can require a large number of iterations of
the $\textit{iterative\_conflict\_analysis}$ procedure. For that reason, it can
be beneficial to use definitions of $\psi$ which return more general coverage
forumlas, resulting in more general coverage constraints that can be computed
in fewer iterations. As the translation of an interpretation is a conjunction,
$\psi(I, e, H, T)$ can be defined as the conjunction of any subset of the
conjuncts of $\mathcal{T}(I, e, T)$, so long as at least one of the conjuncts
does not accept $H$. Definition~\ref{def:psi} presents three such approaches.

\begin{definition}\label{def:psi}
  Let $I$ be an interpretation.

  \begin{itemize}
    \item If $I$ is a model of $H$, $\psi_{\alpha}(I, e, H, T)$ is an arbitrary
      conjunct of $\mathcal{T}_2(I, e, T)$ that does not accept $H$; otherwise,
      $\psi_{\alpha}(I, e, H, T) = \mathcal{T}_1(I, e, T)$.
    \item If $I$ is a model of $H$, $\psi_{\beta}(I, e, H, T) =
      \mathcal{T}_2(I, e, T)$; otherwise, $\psi_{\beta}(I, e, H, T) =
      \mathcal{T}_1(I, e, T)$.
    \item
      $\psi_{\gamma}(I, e, H, T) = \mathcal{T}(I, e, T)$.
  \end{itemize}
\end{definition}

\begin{example}\label{eg:compute_necessary_from_CDPI}
  Reconsider the $\lasne$ task $T$ and the CDPI $e$ from
  Example~\ref{eg:translate}. Let $H = \emptyset$. $H$ does not cover $e$.
  Example executions of the conflict analysis methods for each of the three
  $\psi$'s in Definition~\ref{def:psi} are given below. As interpretations are
  arbitrarily chosen these executions are not unique\footnote{In reality, in
  ILASP4 we employ several heuristics when searching for the $I$'s to try to
  keep the coverage formulas short, meaning that ILASP4 will favour some
  executions over other executions.}.

  \begin{itemize}
    \item First, consider $\psi_{\alpha}$.
      In the first iteration, $F = \bot$, so we need to find an accepting
      answer set for any hypothesis in the space. One such accepting answer set
      is $I_1 = \lbrace \asp{q},$ $\asp{r},$ $\asp{t}\rbrace$ (which is an
      accepting answer set for $\lbrace h^2, h^3\rbrace$). $I_1$ is a model of
      $H$, so $\psi_{\alpha}$ will pick an arbitrary conjunct of
      $\mathcal{T}_2(I_1, e, T) = (h^1_{id} \lor h^3_{id})\land h^2_{id}$ that
      does not accept $H$.
      Let $D_1$ be $(h^1_{id} \lor h^3_{id})$. $F$ becomes $D_1$ at the start
      of the next iteration. At this point, there are no hypotheses that cover
      $e$ that do not accept $F$. Hence, $(h^1_{id} \lor h^3_{id})$ is returned
      as the result of conflict analysis.
    \item
      Next, consider $\psi_{\beta}$.
      Again, let $I_1 = \lbrace \asp{q},$ $\asp{r},$ $\asp{t}\rbrace$ (which is
      an accepting answer set for $\lbrace h^2, h^3\rbrace$). $I_1$ is a model
      of $H$, so $D_1 = \mathcal{T}_2(I_1, e, T) = (h^1_{id} \lor
      h^3_{id})\land h^2_{id}$. $F$ becomes $D_1$ at the start of the next
      iteration.
      Next, let $I_2 = \lbrace \asp{q},$ $\asp{r} \rbrace$ (which is an
      accepting answer set for $\lbrace h^3\rbrace$). $I_2$ is a model of $H$,
      so $D_2 = \mathcal{T}_2(I_2, e, T) = h^3_{id}$. $F$ becomes $D_1\lor D_2$
      at the start of the next iteration.
      At this point, there are no hypotheses that cover $e$ that do not accept
      $F$. Hence, $((h^1_{id} \lor h^3_{id}) \land h^2_{id})\lor h^3_{id}$ is
      returned as the result of conflict analysis.
    \item
      Finally, consider $\psi_{\gamma}$. In the first iteration, $F = \bot$.
      Again, let $I_1 = \lbrace \asp{q},$ $\asp{r},$ $\asp{t}\rbrace$ (which is
      an accepting answer set for $\lbrace h^2, h^3\rbrace$). $D_1 =
      \mathcal{T}(I_1, e, T) = (h^1_{id} \lor h^3_{id})\land h^2_{id}\land
      \lnot h^4_{id}$. $F$ becomes $D_1$ at the start of the next iteration.
      Next, let $I_2 = \lbrace \asp{q},$ $\asp{r} \rbrace$ (which is an
      accepting answer set for $\lbrace h^3\rbrace$). $D_2 = \mathcal{T}(I_2,
      e, T) = h^3_{id} \land \lnot h^2$. $F$ becomes $D_1\lor D_2$ at the start
      of the next iteration.
      Let $I_3 = \lbrace \asp{q},$ $\asp{r},$ $\asp{s},$ $\asp{t}\rbrace$
      (which is an accepting answer set for $\lbrace h^2, h^3, h^4\rbrace$).
      $D_3 = \mathcal{T}(I_3, e, T) = (h^1_{id} \lor h^3_{id}) \land h^3 \land
      h^4_{id}$. $F$ becomes $D_1 \lor D_2 \lor D_3$ at the start of the next
      iteration.
      At this point, there are no hypotheses that cover $e$ that do not accept
      $F$. Hence, $
      ((h^1_{id} \lor h^3_{id}) \land h^2_{id} \land \lnot h^4_{id})
      \lor
      (h^3_{id} \land \lnot h^2)
      \lor
      ((h^1_{id} \lor h^3_{id}) \land h^2_{id} \land h^4_{id})
      $ is returned as the result of conflict analysis.
  \end{itemize}

  This example shows the differences between the different $\psi$'s.
  $\psi_{\gamma}$, used by ILASP3, essentially results in a complete
  translation of the example $e$ -- the coverage formula is satisfied if and
  only if the example is covered. $\psi_{\alpha}$, on the other hand, results
  in a much smaller coverage formula, which is computed in fewer steps, but
  which is only necessary (and not sufficient) for $e$ to be covered.
  Consequently, when $\psi_{\alpha}$ is used, it may be necessary for multiple
  conflict analysis steps on the same example -- for instance, a future
  hypothesis search phase might find $\lbrace h^1\rbrace$, which satisfies the
  coverage formula found using $\psi_{\alpha}$ but does not cover $e$.
  $\psi_{\alpha}$ and $\psi_{\gamma}$ are two extremes: $\psi_{\alpha}$ finds
  very short, easily computable coverage formulas and may require many
  iterations of the CDILP algorithm for each example; and $\psi_{\gamma}$
  finds very long coverage formulas that may take a long time to compute, but
  only requires at most one iteration of the CDILP algorithm per example.
  $\psi_{\beta}$ provides a middle ground. In this example, it finds a formula
  which is equivalent to the one found by $\psi_{\gamma}$ but does so in fewer
  iterations.

\end{example}

The following two theorems show that for each of the three version of $\psi$
presented in Definition~\ref{def:psi}, the iterative conflict analysis
algorithm is guaranteed to terminate and is a valid method for computing a
coverage constraint for a positive CDPI.

\appendixTheorem{cdpi_pos_ca_terminate}{
  Let $e$ be a CDPI and $H\subseteq S_{M}$ be a hypothesis that does not accept
  $e$. For each $\psi \in \lbrace \psi_{\alpha}, \psi_{\beta},
  \psi_{\gamma}\rbrace$, the procedure
  $\textit{iterative\_conflict\_analysis}(T, H, e, \psi)$ is guaranteed to
  terminate and return a coverage formula $F_{\psi}$.  Furthermore, for each
  $\psi$:
  \begin{enumerate}
    \item $F_{\psi}$ does not accept $H$.
    \item If $e$ is a positive example, the pair $\langle e, F_{\psi}\rangle$
      is a coverage constraint.
  \end{enumerate}
}{
  First, we show that in each case, $\textit{iterative\_conflict\_analysis}(T,
  H, e, \psi)$ is guaranteed to terminate. As $F$ is a disjunction, and each
  iteration of the while loop adds an extra disjunct, the hypotheses
  $H'\subseteq S_M$ accepted must either increase or stay the same with each
  iteration. Hence as there are a finite number of hypotheses (because $S_M$ is
  finite), it remains to show that each iteration adds at least one hypothesis
  that is accepted by $F$ at the end of the iteration that was not accepted by
  $F$ at the start of the iteration.

  Next note, that in any iteration the hypothesis $H'$ referenced in the
  condition of the while loop is not accepted by $F$ at the start of the
  iteration. But as $I\in AAS(e, B\cup H')$, $H'$ must be accepted by
  $\mathcal{T}(I, e, T)$ (by Theorem~\ref{thm:translateCDPI}); hence, as for each
  $\psi$ the formulas that are accepted by $\mathcal{T}(I, e, T)$ are accepted
  by $\psi(I, e, H, T)$ (because it is one of the conjuncts of $\mathcal{T}(I,
  e, T)$), $H'$ must be accepted by $F$ at the end of the iteration.
  Hence, in all three cases, $\textit{iterative\_conflict\_analysis}(T, H, e,
  \psi)$ is guaranteed to terminate.

  We now show the remaining points of the theorem.

  \begin{enumerate}
    \item
      It suffices to show that in each iteration, the new disjunct added to $F$
      does not accept $H$. Consider an arbitrary iteration let $I$ be the
      interpretation fixed in this iteration.
      As $I\in AAS(e, B\cup H')$, $I$ cannot be in $AAS(e, B\cup H)$ (or $H$
      would accept $e$). Therefore, by Theorem~\ref{thm:translateCDPI}, $H$ must
      not accept $\mathcal{T}(I, e, T)$. Hence $\psi_{\gamma}(I, e, H, T)$ does
      not accept $H$.

      If $I$ is a model of $H$, then there is no rule $R\in H$ s.t.\ $I$ is not
      a model of $R$. Hence, $\mathcal{T}_{1}(I, e, T)$ must accept $H$,
      meaning that there must be at least one conjunct of $\mathcal{T}_2(I, e,
      T)$ that does not accept $H$. Hence, $\psi_{\alpha}(I, e, H, T)$ is well
      defined and neither $\psi_{\alpha}(T, e, H, T)$ nor $\psi_{\beta}(T, e,
      H, T)$ accept $H$.

      Hence, in all three cases, $F_{\psi}$ does not accept $H$.
    \item
      As the while loop terminates, when $F$ is returned it must be the case
      that the negation of the while condition holds. Hence, there is no $I$
      for which there is an $H'\subseteq S_M$ s.t.\ $F$ does not accept $H'$
      and $I\in AAS(e, B \cup H)$. If $e$ is a positive example, this is
      equivalent to saying that there is no hypothesis that covers $H$ that is
      not accepted by $F_{\psi}$. Hence, every hypothesis that covers $e$ must
      be accepted by $F$, meaning that for each $\psi$, $\langle e,
      F_{\psi}\rangle$ is a coverage constraint.
  \end{enumerate}

}

\subsection{Negative CDPI examples}

This section presents two methods of conflict analysis for a negative CDPI $e$.
The first, used by ILASP3, is to call $iterative\_conflict\_analysis(e, H, T,
\psi_{\gamma})$. As the result of this is guaranteed to return a coverage
formula $F$ that is both necessary and sufficient for $e$ to be accepted, the
negation of this formula ($\lnot F$) is guaranteed to be necessary and
sufficient for $e$ to not be accepted -- i.e.\ for $e$ to be covered (as $e$ is
a negative example). This result is formalised by Theorem~\ref{thm:neg_ilasp3}.

\appendixTheorem{neg_ilasp3}{
  Let $e$ be a negative CDPI and $H$ be a hypothesis that does not cover $e$.
  Then $\textit{iterative\_conflict\_analysis}(e, H, T, \psi_{\gamma})$ is
  guaranteed to terminate, returning a coverage formula $F$. Furthermore, the
  pair $\langle e, \lnot F\rangle$ is a coverage constraint and $H$ is not
  accepted by $\lnot F$.
}{
  (Similar to the proof of Theorem~\ref{thm:cdpi_pos_ca_terminate}).
  As $F$ is a disjunction, and each iteration of the while loop adds an extra
  disjunct, the hypotheses $H'\subseteq S_M$ accepted must either increase or
  stay the same with each iteration. Hence as there are a finite number of
  hypotheses (because $S_M$ is finite), it remains to show that each iteration
  adds at least one hypothesis that is accepted by $F$ at the end of the
  iteration that was not accepted by $F$ at the start of the iteration.

  Next note, that in any iteration the hypothesis $H'$ referenced in the
  condition of the while loop is not accepted by $F$ at the start of the
  iteration. But as $I\in AAS(e, B\cup H')$, $H'$ must be accepted by
  $\mathcal{T}(I, e, T)$ (by Theorem~\ref{thm:translateCDPI}); hence, $H'$ must be
  accepted by $F$ at the end of the iteration.
  Hence, $\textit{iterative\_conflict\_analysis}(T, H, e, \psi)$ is guaranteed
  to terminate.

  To show that $\langle e, \lnot F\rangle$ is a coverage constraint, we must
  show that no hypothesis that covers $e$ is accepted by $F$. We can show this
  be proving that in an arbitrary iteration, no hypothesis that covers $e$ is
  accepted by the new disjunct added to $F$. Each such disjunct is equal to
  $\mathcal{T}(I, e, T)$ for some interpretation $I$ that extends $e_{pi}$. For
  any hypothesis $H'$ that is accepted by this disjunct, $I$ is an accepting
  answer set of $B\cup H'$ w.r.t.\ $e$. Hence, any such $H'$ cannot cover $e$.
  Therefore, no hypothesis that covers $e$ can be accepted by the disjunct.
  Hence, $\langle e, \lnot F\rangle$ is a coverage constraint.

  Finally, we must show that $H$ is not accepted by $F$. Assume for
  contradiction that $H$ is accepted by $F$. As $H$ does not cover $e$, there
  must be an $I\in AAS(B\cup H, e)$, which contradicts the fact that the while
  loop terminated. Hence, $H$ is not accepted by $F$.
}

\begin{example}
  Reconsider the $\lasne$ task $T$ and the CDPI $e$ from
  Example~\ref{eg:translate}, but this time let $e$ be a negative example. Let
  $H = \lbrace h^1, h^2\rbrace$. $H$ does not cover $e$ as $B\cup H \cup
  e_{ctx}$ has an answer set $\lbrace \asp{q},$ $\asp{r},$ $\asp{t}\rbrace$
  that contains $\asp{r}$. As shown in
  Example~\ref{eg:compute_necessary_from_CDPI},
  $\textit{iterative\_conflict\_analysis}(e, H, T, \psi_{\gamma})$ returns the
  formula $F = ((h^1_{id} \lor h^3_{id}) \land h^2_{id} \land \lnot h^4_{id})
  \lor (h^3_{id} \land \lnot h^2) \lor ((h^1_{id} \lor h^3_{id}) \land h^2_{id}
  \land h^4_{id})$. $F$ is accepted by exactly those hypotheses which accept
  $e$; hence, $\lnot F$ is satisfied by exactly those hypotheses which do not
  accept $e$ (i.e.\ those hypotheses which cover $e$). Hence, $\langle e, \lnot
  F\rangle$ is a coverage constraint.
\end{example}

The method for computing a necessary constraint for a negative CDPI $e$ is much
simpler than for positive CDPIs. Note that each disjunct in the formula $F$
computed by $\textit{iterative\_conflict\_analysis}(e, H, T, \psi_{\gamma})$ is
sufficient but not necessary for the CDPI $e$ to be accepted (i.e.\ for $e$ to
not be covered, as it is a negative example), and therefore its negation is
necessary but not sufficient for $e$ to be covered. This means that to compute
a necessary constraint for a negative CDPI, we only need to consider a single
interpretation. This is formalised by the following theorem. The approach used
in ILASP4 computes an arbitrary such coverage formula. Note that this is
guaranteed to terminate and the following theorem shows that the method is a
valid method for conflict analysis.

\appendixTheorem{neg_ilasp4}{
  Let $e$ be a CDPI in $E^{-}$ and $H\subseteq S_{M}$ be a hypothesis that does
  not cover $e$. $\textit{AAS}(e, B\cup H)$ is non-empty and for any $I \in
  \textit{AAS}(e, B\cup H)$:
  \begin{enumerate}
    \item $\lnot \mathcal{T}(I, e, T)$ does not accept $H$.
    \item $\langle e, \lnot \mathcal{T}(I, e, T)\rangle$ is a coverage constraint.
  \end{enumerate}
}{
  As $H$ does not cover $e$, $\textit{AAS}(e, B\cup H)$ must be non-empty (by
  definition of a negative example).
  \begin{enumerate}
    \item Let $I$ be an arbitrary interpretation $I \in \textit{AAS}(e, B\cup
      H)$. By Theorem~\ref{thm:translateCDPI}, $\mathcal{T}(I, e, T)$ accepts
      $H$. Hence, $\lnot \mathcal{T}(I, e, T)$ does not accept $H$.
    \item
      Let $H'$ be a hypothesis that covers $e$. It suffices to show that $H'$
      is accepted by $\lnot \mathcal{T}(I, e, T)$. As $H'$ covers $e$, $I$
      cannot be an answer set of $B\cup e_{ctx} \cup H'$. Hence, by
      Theorem~\ref{thm:translateCDPI}, $\mathcal{T}(I, e, T)$ does not accept
      $H'$; hence $\lnot \mathcal{T}(I, e, T)$ accepts $H$.
  \end{enumerate}
}

\begin{example}
  Again, reconsider the $\lasne$ task $T$ and the CDPI $e$ from
  Example~\ref{eg:translate}, letting $e$ be a negative example. Let $H =
  \lbrace h^1, h^2\rbrace$, which does not cover $e$. $I_1 = \lbrace \asp{q}$,
  $\asp{t}$, $\asp{r}\rbrace \in AAS(e, B \cup H)$. $\mathcal{T}(I_1, e, T) =
  \lnot h^4_{id}\land (h^1_{id}\lor h^3_{id})\land h^2_{id}$. Clearly, $H$ is
  not accepted by $\lnot \mathcal{T}(I_1, e, T)$.

  As shown by Theorem~\ref{thm:translateCDPI}, $\mathcal{T}(I_1, e, T)$ is
  accepted by exactly those hypotheses $H'$ s.t.\ $I_1 \in AAS(e, B\cup H')$.
  Hence, any hypothesis that accepts $F$ does not cover $e$. Therefore $\langle
  e, \lnot \mathcal{T}(I_1, e, T)\rangle$ is a coverage constraint.
\end{example}

\subsection{ILASP3 and ILASP4}

ILASP3 and ILASP4 are both instances of the CDILP approach to ILP formalised in
this paper. The difference between the two algorithms is their approaches to
conflict analysis.

ILASP3 uses $\psi_{\gamma}$ for conflict analsysis on positive examples and
essentially uses the same approach for negative examples, negating the
resulting coverage formula (Theorem~\ref{thm:neg_ilasp3} proves that this is a
valid method for conflict analysis). Hence, by Theorem~\ref{thm:sac}, ILASP3 is
sound and complete and is guaranteed to terminate on any well-formed learning
task.

The conflict analysis methods adopted by ILASP3 may result in extremely long
coverage constraints that take a very long time to compute.  This is more
apparent on some learning problems than others. The following definition
defines two classes of learning task: \emph{categorical} learning tasks for
which all programs that need to be considered to solve the task have a single
answer set; and \emph{non-categorical} learning tasks, in which some programs
have multiple answer sets. The performance of ILASP3 and ILASP4 on categorical
learning tasks is likely to be very similar, whereas on non-categorical tasks,
ILASP3's tendancy to compute long coverage constraints is more likely to be an
issue, because larger numbers of answer sets tend to lead to longer coverage
constraints.

\begin{definition}
We say that a learning task $T$ is \emph{categorical} if for each CDPI $e$ in
$T$, there is at most one interpretation $I$ such that there is a hypothesis
$H\subseteq S_M$ s.t.\ $I\in AAS(e, B\cup H)$. A learning task which is not
categorical is called \emph{non-categorical}.
\end{definition}

In the evaluation in the next section, we consider two versions of ILASP4. The
first (ILASP4a) uses $\psi_{\alpha}$ and the coverage constraints for negative
examples defined in Theorem~\ref{thm:neg_ilasp4}. The second (ILASP4b) uses
$\psi_{\beta}$ and the same notion of coverage constraints for negative
examples. As we have shown that these approaches to conflict analysis are
valid, both ILASP4a and ILASP4b are sound and complete and are guaranteed to
terminate on any well-formed learning task. Both approaches produce more
general coverage constraints than ILASP3 (with ILASP4b being a middle ground
between ILASP4a and ILASP3), meaning that each iteration of the CDILP approach
is likely to be faster (on non-categorical tasks). The trade-off is that
whereas in ILASP3 the coverage constraint computed for an example $e$ is
specific enough to rule out any hypothesis that does not cover $e$, meaning
that each example will be processed at most once, in ILASP4, this is not the
case and each example may be processed multiple times. So ILASP4 may have a
larger number of iterations of the CDILP procedure than ILASP3; however,
because each of these iterations is likely to be shorter, in practice, ILASP4
tends to be faster overall. This is supported by the experimental results in
the next section.

\section{Evaluation}
\label{sec:eval}
This section presents an evaluation of ILASP's conflict-driven approach to ILP.
The datasets used in this paper have been previously used to evaluate previous
versions of ILASP, including ILASP3, (e.g.\ in~\cite{ILASP_thesis,ACS18}).
ILASP3 has previously been applied to several real-world datasets, including
event detection, sentence chunking and preference learning~\cite{ACS18}. Rather
than repeating these experiments here, we direct the reader to~\cite{ACS18},
which also gives a detailed comparison between the performance of ILASP3 and
other ILP systems on noisy datasets. In this evaluation, we focus on synthetic
datasets which highlight the weaknesses of older ILASP systems, and show how
(in particular) ILASP4 has overcome them.

\subsection{Comparison between ILASP versions on benchmark tasks}

In~\cite{ICLP16}, ILASP was evaluated on a set of non-noisy benchmark problems,
designed to test all functionalities of ILASP at the time (at that time, ILASP
was incapable of solving noisy learning tasks). The running times of all
incremental versions of ILASP (ILASP1 and ILASP2 are incapable of solving large
tasks) on two of these benchmarks are shown in
Table~\ref{tbl:benchmarks}.\footnote{
  All experiments in this paper were run on an Ubuntu 20.04 virtual machine
  with 8 cores and 16GB of RAM, hosted on a server with a 3.0GHz
  Intel$^{\tiny{\textregistered}}$ Xeon$^{\tiny{\textregistered}}$ Gold 6136
  processor, unless otherwise noted. All benchmark tasks in this section are
  available for download from \url{http://www.ilasp.com/research}.
} The remaining results are available in Appendix A, showing how the different
versions of ILASP compare on weak constraint learning tasks.

\begin{table}
  \footnotesize
  \begin{tabular}{c ccccc ccccc}
    \hline
    Task   & $|S_M|$ & $|E^{+}|$ & $|E^{-}|$ & $|O^{b}|$ & $|O^{c}|$ & 2i & 3 & 4a & 4b \\\hline\hline
    Hamilton                & 104       & 100 & 100 & 0   & 0   & 3.09  & 64.12 & 4.51  & 2.59 \\
    Noisy Hamilton A        & 104       & 29  & 31  & 0   & 0   & 237.12& 35.67 & 8.98  & 5.78 \\
    Noisy Hamilton B        & 104       & 56  & 64  & 0   & 0   & TO    & 68.31 & 25.11 & 21.57 \\
    Noisy Hamilton C        & 104       & 87  & 93  & 0   & 0   & TO    & 62.98 & 46.95 & 43.59 \\\hline
    Agent A                 & 531       & 200 & 0   & 0   & 0   & 29.39 & 57.34 & 42.36 & 32.86\\
    Agent B                 & 146       & 50  & 0   & 0   & 0   & 2.56  & 1091.98 & 15.22 & 11.97\\
    Agent C                 & 160       & 80  & 120 & 0   & 0   & 14.24 & 14.73 & 12.43 & 9.27 \\
  \end{tabular}
  \caption{The running times (in seconds) of various ILASP systems on the set
  of benchmark problems. TO denotes a timeout (where the time limit was
  1800s).\label{tbl:benchmarks}}
\end{table}

The first benchmark problem is to learn the definition of whether a graph is
Hamiltonian or not (i.e.\ whether it contains a Hamilton cycle). The background
knowledge is empty and each example corresponds to exactly one graph,
specifying which $\asp{node}$ and $\asp{edge}$ atoms should be true. Positive
examples correspond to Hamiltonian graphs, and negative examples correspond to
non-Hamiltonian graphs. This is the context-dependent ``Hamilton B'' setting
from~\cite{ICLP16}. ILASP2i and both versions of ILASP4 perform similarly, but
ILASP3 is significantly slower than the other systems. This is because the
Hamiltonian learning task is non-categorical and the coverage formulas
generated by ILASP3 tend to be large. This experiment was repeated with a
``noisy'' version of the problem where 5\% of the examples were mislabelled
(i.e.\ positive examples were changed to negative examples or negative examples
were changed to positive examples). To show the scalability issues with ILASP2i
on noisy learning tasks, three versions of the problem were run, with 60, 120
and 180 examples. ILASP2i's execution time rises rapidly as the number of
examples grows, and it is unable to solve the last two tasks within the time
limit of 30 minutes. ILASP3 and ILASP4 are all able to solve every version of
the task in far less than the time limit, with ILASP4b performing best. The
remaining benchmarks are drawn from non-noisy datasets, where ILASP2i performs
fairly well.

The second setting originates from~\cite{JELIAILASP} and is based on an agent
learning the rules of how it is allowed to move within a grid. Agent A requires
a hypothesis describing the concept of which moves are valid, given a history
of where an agent has been. Examples are of the agent's history of moving
through the map and a subset of the moves which were valid/invalid at each time
point in its history. Agent B requires a similar hypothesis to be learned, but
with the added complexity that an additional concept is required to be
invented (and used in the rest of the hypothesis). In Agent C, the hypothesis
from Agent A must be learned along with a constraint ruling out histories in
which the agent visits a cell twice (not changing the definition of valid
move). This requires negative examples to be given, in addition to positive
examples. Although scenarios A and C are technically
non-categorical, scenario B causes more of an issue for ILASP3 because of the
(related) challenge of predicate invention. The potential to invent new
predicates which are unconstrained by the examples means that there are many
possible answer sets for each example, which leads ILASP3 to generate extremely
long coverage formulas. In this case ILASP3 is nearly two orders of magnitude
slower than either version of ILASP4. As the Agent tasks are non-noisy and have a
relatively small problem domain, ILASP2i solves these task
fairly easily and, in the case of Agent A and Agent B, in less time than either
version of ILASP4. On simple non-noisy tasks, the computation of coverage
constraints are a computational overhead that can take longer than using a
meta-level approach such as ILASP2i.

\subsection{Comparison between methods for conflict analysis on a synthetic noisy dataset}

In~\cite{ACS18} ILASP3 was evaluated on a synthetic noisy dataset in which the
task is to learn the definition of what it means for a graph to be Hamiltonian.
This concept requires learning a hypothesis that contains choice rules,
recursive rules and hard constraints, and also contains negation as failure.
The advantage of using a synthetic dataset is that the amount of noise in the
dataset (i.e.\ the number of examples which are mislabelled) can be controlled
when constructing the dataset. This allows us to evaluate ILASP's tolerance to
varying amounts of noise. ILASP1, ILASP2 and ILASP2i (although theoretically
capable of solving any learning task which can be solved by the later systems)
are all incapable of solving large noisy learning tasks in a reasonable amount
of time.  Therefore, this section only presents a comparison of the performance
of ILASP3 and ILASP4 on the synthetic noisy dataset from~\cite{ACS18}.

\begin{figure*}[t]

    \hfill
        \includegraphics[width=0.32\textwidth, trim={15mm 0mm 17mm 0mm},clip]{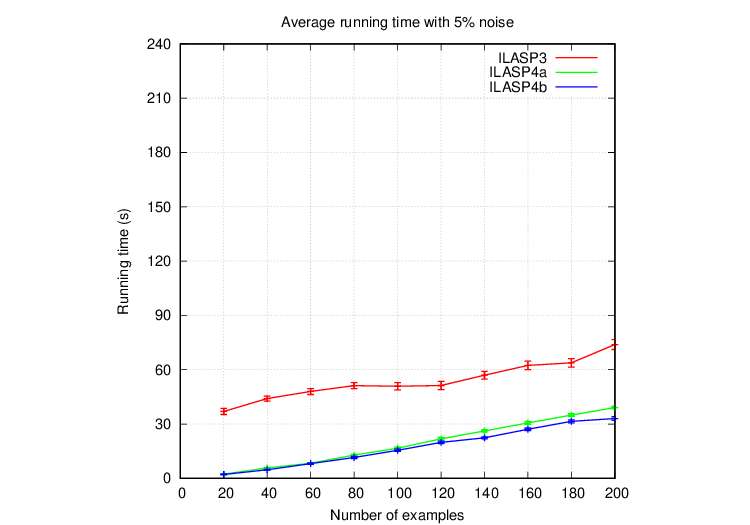}
        \hfill
        \includegraphics[width=0.32\textwidth, trim={15mm 0mm 17mm 0mm},clip]{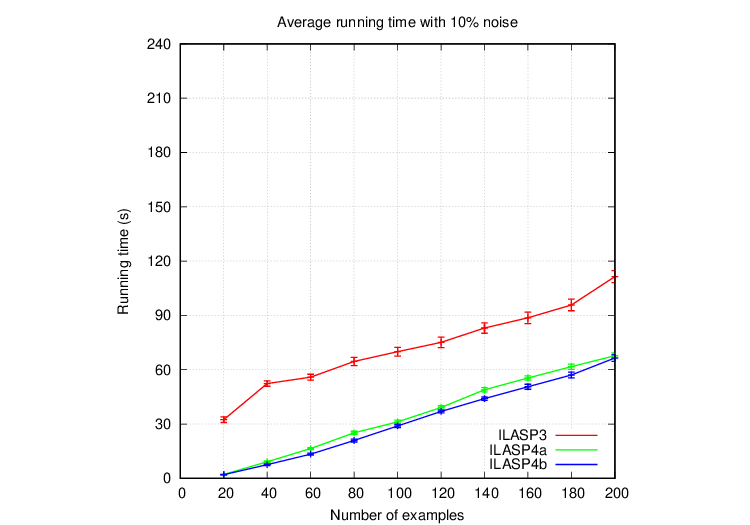}
        \hfill
        \includegraphics[width=0.32\textwidth, trim={15mm 0mm 17mm 0mm},clip]{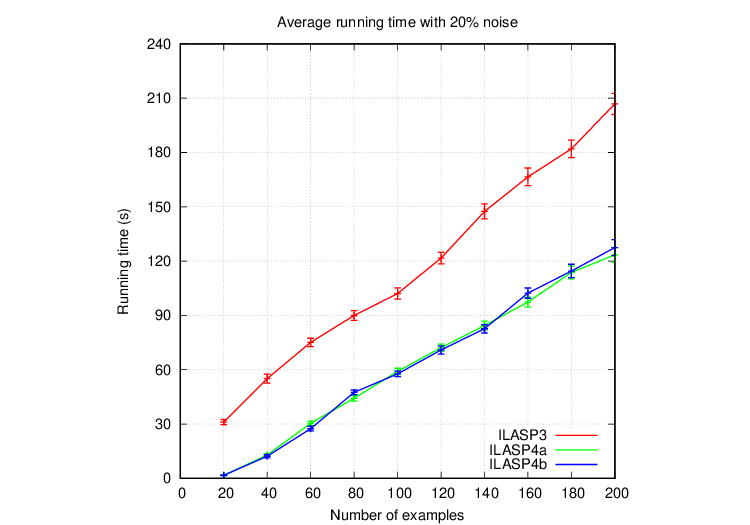}
        \hfill

    \caption{\label{fig:hamilton} The average computation time of ILASP3,
    ILASP4a and ILASP4b for the Hamilton learning task, with varying numbers of
    examples, with 5, 10 and 20\% noise.}

\end{figure*}

\begin{figure*}[t]

    \hfill
        \includegraphics[width=0.32\textwidth, trim={15mm 0mm 17mm 0mm},clip]{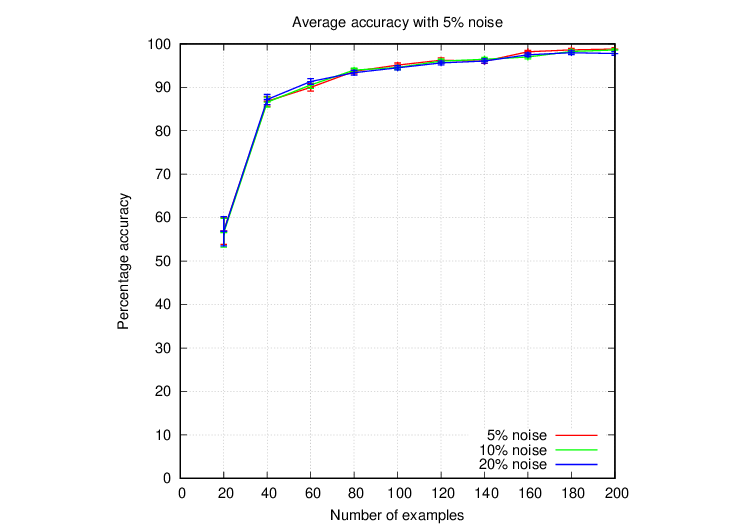}
        \hfill
        \includegraphics[width=0.32\textwidth, trim={15mm 0mm 17mm 0mm},clip]{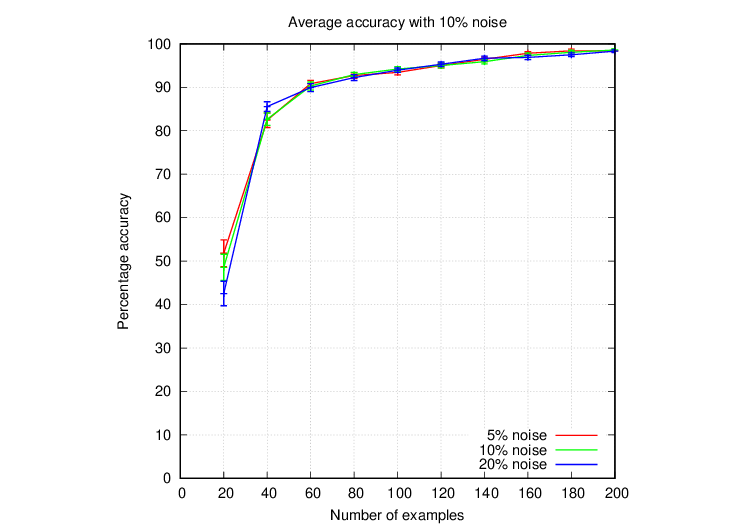}
        \hfill
        \includegraphics[width=0.32\textwidth, trim={15mm 0mm 17mm 0mm},clip]{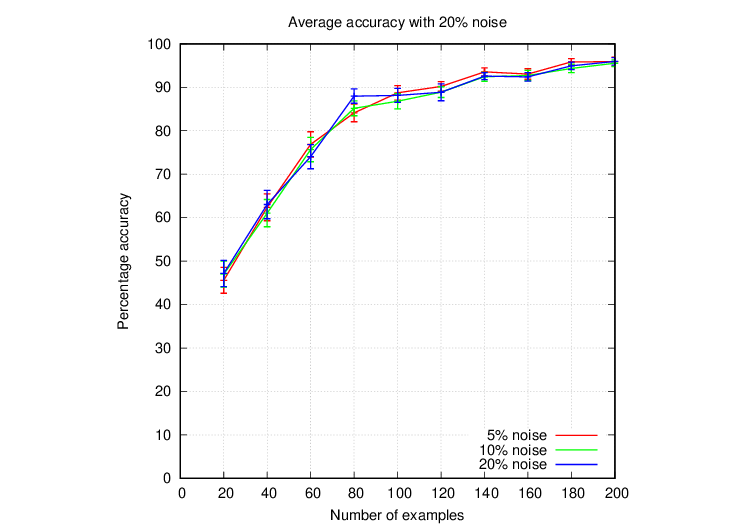}
        \hfill

    \caption{\label{fig:hamilton_acc} The average accuracies of ILASP3, ILASP4a
    and ILASP4b for the Hamilton learning task, with varying numbers of
    examples, with 5, 10 and 20\% noise.}

\end{figure*}

For $n = 20, 40, \ldots, 200$, $n$ random graphs of size one to four were
generated, half of which were Hamiltonian. The graphs were labelled as either
positive or negative, where positive indicates that the graph is Hamiltonian.
Three sets of experiments were run, evaluating each ILASP algorithm with 5\%,
10\% and 20\% of the examples being labelled incorrectly. In each experiment,
an equal number of Hamiltonian graphs and non-Hamiltonian graphs were randomly
generated and 5\%, 10\% or 20\% of the examples were chosen at random to be
labelled incorrectly. This set of examples were labelled as positive (resp.\
negative) if the graph was not (resp.\ was) Hamiltonian. The remaining examples
were labelled correctly (positive if the graph was Hamiltonian; negative if the
graph was not Hamiltonian). Figures~\ref{fig:hamilton}
and~\ref{fig:hamilton_acc} show the average running time and accuracy
(respectively) of each ILASP version with up to 200 example graphs. Each
experiment was repeated 50 times (with different randomly generated examples).
In each case, the accuracy was tested by generating a further 1,000 graphs and
using the learned hypothesis to classify the graphs as either Hamiltonian or
non-Hamiltonian.

The experiments show that each of the three conflict-driven ILASP algorithms
(ILASP3, ILASP4a and ILASP4b) achieve the same accuracy on average (this is to
be expected, as each system is guaranteed to find an optimal solution of any
task). They each achieve a high accuracy (of well over 90\%), even with 20\% of
the examples labelled incorrectly. A larger percentage of noise means that
ILASP requires a larger number of examples to achieve a high accuracy. This is
to be expected, as with few examples, the hypothesis is more likely to
``overfit'' to the noise, or pay the penalty of some non-noisy examples. With
large numbers of examples, it is more likely that ignoring some non-noisy
examples would mean not covering others, and thus paying a larger penalty. The
computation time of each algorithm rises in all three graphs as the number of
examples increases. This is because larger numbers of examples are likely to
require larger numbers of iterations of the CDILP approach (for each ILASP
algorithm). Similarly, more noise is also likely to mean a larger number of
iterations. The experiments also show that on average, both ILASP4 approaches
perform around the same, with ILASP4b being marginally better than ILASP4a.
Both ILASP4 approaches perform significantly better than ILASP3. Note that the
results reported for ILASP3 on this experiment are significantly better than
those reported in~\cite{ACS18}. This is due to improvements to the overall
ILASP implementation (shared by ILASP3 and ILASP4).

\begin{figure*}[t]

  \centering

        \includegraphics[width=0.40\textwidth, trim={15mm 0mm 17mm 0mm},clip]{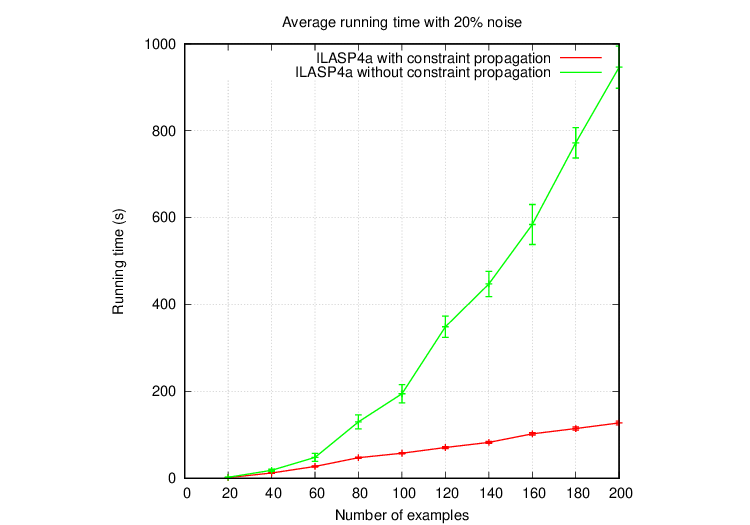}

    \caption{\label{fig:cp} The average running times of ILASP4a with and
    without constraint propagation enabled for the Hamilton learning task with
    20\% noise.}

\end{figure*}

\paragraph{The effect of constraint propagation.}
The final experiment in this section evaluates the benefit of using constraint
propagation on noisy learning tasks. The idea of constraint propagation is that
although it itself takes additional time, it may decrease the number of
iterations of the conflict-driven algorithms, meaning that the overall running
time is reduced. Figure~\ref{fig:cp} shows the difference in running times
between ILASP4a with and without constraint propagation enabled on a repeat of
the Hamilton 20\% noise experiment. Constraint propagation makes a huge
difference to the running times demonstrating that this feature of CDILP is a
crucial factor in ILASP's scalability over large numbers of noisy examples.

\section{Related Work}
\label{sec:related}
\paragraph{Learning under the answer set semantics.}
Traditional approaches to learning under the answer set semantics were broadly
split into two categories: \emph{brave} learners (e.g.\
\cite{Sakama2009,ray2009nonmonotonic,corapi2012,ILED,inspire}), which aimed to
explain a set of (atomic) examples in at least one answer set of the learned
program; and \emph{cautious} learners (e.g.\
\cite{inoue1997learning,seitzer2000inded,sakama2000inverse,Sakama2009}), which
aimed to explain a set of (atomic) examples in every answer set of the learned
program.\footnote{Some of these systems predate the terms brave and cautious
induction, which first appeared in~\cite{Sakama2009}.} In general, it is not
possible to \emph{distinguish} between two ASP programs (even if they are not
strongly equivalent) using either brave or cautious reasoning
alone~\cite{AIJ17}, meaning that some programs cannot be learned with either
brave or cautious induction; for example, no brave induction system is capable
of learning constraints -- roughly speaking, this is because examples in brave
induction only say what \emph{should} be (in) an answer set, so can only
incentivise learning programs with new or modified answer sets (compared to the
background knowledge on its own), whereas constraints only rule out answer
sets.  ILASP~\cite{JELIAILASP} was the first system capable of combining brave
and cautious reasoning, and (resources permitting) can learn any ASP
program\footnote{Note that some ASP constructs, such as aggregates in the
bodies of rules, are not yet supported by the implementation of ILASP, but the
abstract algorithms are all capable of learning them.} up to strong equivalence
in ASP~\cite{AIJ17}.

FastLAS~\cite{lawfastlas} is a recent ILP system that solves a restricted
version of ILASP's learning task. Unlike ILASP, it does not enumerate the
hypothesis space in full, meaning that it can scale to solve tasks with much
larger hypothesis spaces than ILASP. Although FastLAS has recently been
extended~\cite{fastlas2}, the restrictions on the extended version still mean
that FastLAS is currently incapable of learning recursive definitions,
performing predicate invention, or of learning weak constraints. Compared to
ILASP, these are major restrictions, and work to lift them is ongoing.

\paragraph{Conflict-driven solvers.}

ILASP's CDILP approach was partially inspired by conflict-driven
SAT~\cite{lynce2003effect} and
ASP~\cite{gebser2007conflict,gekakaosscsc11a,alviano2013wasp} solvers, which
generate \emph{nogoods} or \emph{learned constraints} (where the term learned
should not be confused with the notion of learning in this paper) throughout
their execution. These nogoods/learned constraints are essentially reasons why
a particular search branch has failed, and allow the solver to rule out any
further candidate solutions which fail for the same reason. The coverage
formulas in ILASP perform the same function. They are a reason why the most
recent hypothesis is not a solution (or, in the case of noisy learning tasks,
not as good a solution as it was previously thought to be) and allow ILASP to
rule out (or, in the case of noisy learning tasks, penalise) any hypothesis
that is not accepted by the coverage formula.

It should be noted that although ILASP3 and ILASP4 are the closest linked ILASP
systems to these conflict-driven solvers, earlier ILASP systems are also
partially conflict-driven. ILASP2~\cite{ICLP15} uses a notion of a
\emph{violating reason} to explain why a particular negative example is not
covered. A violating reason is an accepting answer set of that example (w.r.t.\
$B\cup H$). Once a violating reason has been found, not only the current
hypothesis, but any hypothesis which shares this violating reason is ruled out.
ILASP2i~\cite{ICLP16} collects a set of \emph{relevant examples} -- a set of
examples which were not covered by previous hypotheses -- which must be covered
by any future hypothesis. However, these older ILASP systems do not extract
coverage formulas from the violating reasons/relevant examples, and use an
expensive meta-level ASP representation which grows rapidly as the number of
violating reasons/relevant examples increases. They also do not have any notion
of \emph{constraint propagation}, which is crucial for efficient solving of
noisy learning tasks.

\paragraph{Incremental approaches to ILP.}
Some older ILP systems, such as ALEPH~\cite{srinivasan2001aleph},
Progol~\cite{muggleton1995inverse} and HAIL~\cite{ray2003hybrid}, incrementally
consider each positive example in turn, employing a \emph{cover loop}. The idea
behind a cover loop is that the algorithm starts with an empty hypothesis $H$,
and in each iteration adds new rules to $H$ such that a single positive example
$e$ is covered, and none of the negative examples are covered. Unfortunately,
cover loops do not work in a non-monotonic setting because the examples covered
in one iteration can be ``uncovered'' by a later iteration. Worse still, the
wrong choice of hypothesis in an early iteration can make another positive
example impossible to cover in a later iteration. For this reason, most ILP
systems under the answer set semantics (including ILASP1 and ILASP2) tend to be
\emph{batch learners}, which consider all examples at once. The CDILP approach
in this paper does not attempt to learn a hypothesis incrementally (the
hypothesis search starts from scratch in each iteration), but instead builds
the set of coverage constraints incrementally. This allows ILASP to avoid the
problems of cover loop approaches in a non-monotonic setting, while still
overcoming the scalability issues associated with batch learners.

There are two other incremental approaches to ILP under the answer set
semantics. ILED~\cite{ILED}, is an incremental version of the XHAIL algorithm,
which is specifically targeted at learning Event Calculus theories. ILED's
examples are split into \emph{windows}, and ILED incrementally computes a
hypothesis through \emph{theory revision}~\cite{wrobel1996first} to cover the
examples. In an arbitrary iteration, ILED revises the previous hypothesis $H$
(which is guaranteed to cover the first $n$ examples), to ensure that it covers
the first $n+1$ examples. As the final hypothesis is the outcome of the series
of revisions, although each revision may have been optimal, ILED may terminate
with a sub-optimal inductive solution. In contrast, every version of ILASP will
always terminate with an optimal inductive solution if one exists.
The other incremental ILP system under the answer set semantics is
RASPAL~\cite{raspal,raspal_thesis}, which uses an ASPAL-like~\cite{corapi2012}
approach to iteratively revise a hypothesis until it is an optimal inductive
solution of a task. RASPAL's incremental approach is successful as it often
only needs to consider small parts of the hypothesis space, rather than the
full hypothesis space. Unlike ILED and ILASP, however, RASPAL considers the
full set of examples when searching for a hypothesis.

Popper~\cite{popper} is a recent approach to learning definite programs. It is
closely related to CDILP as it also uses an iterative approach where the
current hypothesis (if it is not a solution) is used to constrain the future
search. However, unlike ILASP, Popper does not extract a coverage formula from
the current hypothesis and counterexample, but instead uses the hypothesis
itself as a constraint; for example, ruling out any hypothesis that
theta-subsumes the current hypothesis. Popper's approach has the advantage
that, unlike ILASP, it does not need to enumerate the hypothesis space in full;
however, compared to ILASP it is very limited, and does not support negation as
failure, choice rules, disjunction, hard or weak constraints, non-observational
predicate learning, predicate invention or learning from noisy examples. It is
unclear whether the approach of Popper could be extended to overcome these
limitations.

\paragraph{ILP approaches to noise.}

Most ILP systems have been designed for the task of learning from example
atoms. In order to search for best hypotheses, such systems normally use a
scoring function, defined in terms of the coverage of the examples and the
length of the hypothesis (e.g.\ ALEPH~\cite{srinivasan2001aleph},
Progol~\cite{muggleton1995inverse}, and the implementation of
XHAIL~\cite{bragaglia2015nonmonotonic}). When examples are noisy, this scoring
function is sometimes combined with a notion of maximum threshold, and the
search is not for an optimal solution that minimises the number of uncovered
examples, but for a hypothesis that does not fail to cover more than a defined
maximum threshold number of examples (e.g.\
\cite{srinivasan2001aleph,hyper_n,raspal}).  In this way, once an acceptable
hypothesis (i.e.\ a hypothesis that covers a sufficient number of examples) is
computed the system does not search for a better one. As such, the
computational task is simpler, and therefore the time needed to compute a
hypothesis is shorter, but the learned hypothesis is not optimal. Furthermore,
to guess the ``correct'' maximum threshold requires some idea of how much noise
there is in the given set of examples. For instance, one of the inputs to the
HYPER/N~\cite{hyper_n} system is the proportion of noise in the examples. When
the proportion of noise is unknown, too small a threshold could result in the
learning task being unsatisfiable, or in learning a hypothesis that overfits
the data. On the other hand, too high a threshold could result in poor
hypothesis accuracy, as the hypothesis may not cover many of the examples. The
$ILP_{LAS}^{noise}$ framework addresses the problem of computing optimal
solutions and in doing so does not require any knowledge a priori of the level
of noise in the data.

Another difference when compared to many ILP approaches that support noise is
that $\lasne$ examples contain partial interpretations. In this paper, we do
not consider penalising individual atoms within these partial interpretations.
This is somewhat similar to what traditional ILP approaches do (it is only the
notion of examples that is different in the two approaches). In fact, while
penalising individual atoms within partial interpretations would certainly be
an interesting avenue for future work, this could be seen as analogous to
penalising the arguments of atomic examples in traditional ILP
approaches~\cite{ILASP_thesis}.

XHAIL is a brave induction system that avoids the need to enumerate the entire
hypothesis space. XHAIL has three phases: abduction, deduction and induction.
In the first phase, XHAIL uses abduction to find a minimal subset of some
specified ground atoms. These atoms, or a generalisation of them, will appear
in the head of some rule in the hypothesis. The deduction phase determines the
set of ground literals which could be added to the body of the rules in the
hypothesis. The set of ground rules constructed from these head and body
literals is called a kernel set. The final induction phase is used to find a
hypothesis which is a generalisation of a subset of the kernel set that proves
the examples. The public implementation of
XHAIL~\cite{bragaglia2015nonmonotonic} has been extended to handle noise by
setting penalties for the examples similarly to $\lasne$. However, as shown in
Example~\ref{eg:XHAIL} XHAIL is not guaranteed to find an optimal inductive
solution of a task.

\begin{example}\label{eg:XHAIL}

Consider the following noisy task, in the XHAIL input format:

\begin{multicols}{2}
\begin{verbatim}
p(X) :- q(X, 1), q(X, 2).
p(X) :- r(X).
s(a).   s(b).   s2(b).
t(1).   t(2).
#modeh r(+s).
#modeh q(+s2, +t).
#example not p(a)=50.
#example p(b)=100.
\end{verbatim}
\end{multicols}

This corresponds to a hypothesis space that contains two facts $F_1$ =
$\asp{r(X)}$, $F_2$ = $\asp{q(X, Y)}$ (in XHAIL, these facts are implicitly
``typed'', so the first fact, for example, can be thought of as the rule
$\asp{r(X) \codeif s(X)}$). The two examples have penalties 50 and 100
respectively. There are four possible hypotheses: $\emptyset$, $F_1$, $F_2$ and
$F_1\cup F_2$, with scores 100, 51, 1 and 52 respectively.  XHAIL terminates
and returns $F_1$, which is a suboptimal hypothesis.

The issue is with the first step. The system finds the smallest abductive
solution, $\lbrace \asp{r(b)}\rbrace$ and as there are no body declarations in
the task, the kernel set contains only one rule: $\asp{r(b) \codeif
s(b)\ruleend}$ XHAIL then attempts to generalise to a first order hypothesis
that covers the examples. There are two hypotheses which are subsets of a
generalisation of $\asp{r(b)}$ ($F_1$ and $\emptyset$); as $F_1$ has a lower score
than $\emptyset$, XHAIL terminates and returns $F_1$.  The system does not find
the abductive solution $\lbrace \asp{q(b, 1)}, \asp{q(b, 2)}\rbrace$, which is
larger than $\lbrace \asp{r(b)}\rbrace$ and is therefore not chosen, even
though it would eventually lead to a better solution than $\lbrace
\asp{r(b)}\rbrace$.

  It should be noted that XHAIL does have an \emph{iterative deepening} feature
  for exploring non-minimal abductive solutions, but in this case using this option
  XHAIL still returns $F_1$, even though $F_2$ is a more optimal hypothesis.
  Even when iterative deepening is enabled, XHAIL only considers non-minimal
  abductive solutions if the minimal abductive solutions do not lead to any
  non-empty inductive solutions.

\end{example}

In comparison to ILASP, in some problem domains, XHAIL is more scalable as it
does not start by enumerating the hypothesis space in full. On the other hand,
as shown by Example~\ref{eg:XHAIL}, XHAIL is not guaranteed to find the optimal
hypothesis, whereas ILASP is. ILASP also solves $ILP_{LAS}^{noise}$ tasks,
whereas XHAIL solves brave induction tasks, which means that due to the
generality results in~\cite{AIJ17} ILASP is capable of learning programs which
are out of reach for XHAIL no matter what examples are given.

Inspire~\cite{inspire} is an ILP system based on XHAIL, but with some
modifications to aid scalability. The main modification is that some rules are
``pruned'' from the kernel set before XHAIL's inductive phase. Both XHAIL and
Inspire use a meta-level ASP program to perform the inductive phase, and the
ground kernel set is generalised into a first order kernel set (using the mode
declarations to determine which arguments of which predicates should become
variables). Inspire prunes rules which have fewer than $Pr$ instances in the
ground kernel set (where $Pr$ is a parameter of Inspire). The intuition is that
if a rule is necessary to cover many examples then it is likely to have many
ground instances in the kernel. Clearly this is an approximation, so Inspire is
not guaranteed to find the optimal hypothesis in the inductive phase. In fact,
as XHAIL is not guaranteed to find the optimal inductive solution of the task
(as it may pick the ``wrong'' abductive solution), this means that Inspire may
be even further from the optimal. The evaluation in~\cite{ACS18} demonstrates
that on a real dataset, Inspire's approximation leads to lower quality
solutions (in terms of the $F_1$ score on a test set) than the optimal
solutions found by ILASP.

\section{Conclusion}

This paper has presented the Conflict-driven Inductive Logic Programming
(CDILP) approach. While the four phases of the CDILP approach are clearly
defined at an abstract level, there is a large range of algorithms that could
be used for the \emph{conflict analysis} phase. This paper has presented two
(extreme) approaches to conflict analysis: the first (used by the
ILASP3 system) extracts as much as possible from a counterexample, computing a
coverage formula which is accepted by a hypothesis if and only if the
hypothesis covers the counterexample; the second (used by the ILASP4 system)
extracts much less information from the example and essentially computes an
explanation as to why the most recent hypothesis does not cover the
counterexample. A third (middle ground) approach is also presented. Our
evaluation shows that the selection of conflict analysis approach is crucial to
the performance of the system and that although the second and third approaches
used by ILASP4 may result in more iterations of the CDILP process than in
ILASP3, because each iteration tends to be much shorter, both versions of
ILASP4 can significantly outperform ILASP3, especially for a particular type of
\emph{non-categorical} learning task.

The evaluation has demonstrated that the CDILP approach is robust to high
proportions of noisy examples in a learning task, and that the constraint
propagation phase of CDILP is crucial to achieving this robustness. Constraint
propagation allows ILASP to essentially ``boost'' the penalty associated with
ignoring a coverage constraint, by expressing that not only the counterexample
associated with the coverage constraint will be left uncovered, but also every
example to which the constraint has been propagated.

There is still much scope for improvement, and future work on ILASP will
include developing new (possibly domain-dependent) approaches to conflict
analysis. The new PyLASP feature of ILASP4 also allows users to potentially
implement customised approaches to conflict analysis, by injecting a Python
implementation of their conflict analysis method into ILASP.

Another avenue of future work is to develop a version of ILASP that does not
rely on computing the hypothesis space before beginning the CDILP process. The
FastLAS~\cite{lawfastlas,fastlas2} systems solve a restricted
$ILP_{LAS}^{noise}$ task and are able to use the examples to compute a small
subset of the hypothesis space that is guaranteed to contain at least one
optimal solution. For this reason, FastLAS has been shown to be far more
scalable than ILASP w.r.t.\ the size of the hypothesis space. However, FastLAS
is far less general than ILASP, reducing its applicability. In future work, we
aim to unify the two lines of research and produce a (conflict driven) version
of ILASP that uses techniques based on FastLAS to avoid needing to compute the
entire hypothesis space.

\bibliographystyle{acmtrans}
\bibliography{paper}
%

\section*{Appendix A: Conflict-driven Learning of Weak Constraints}
In addition to the rules considered in the paper, ILASP is able to learn weak
constraints~\cite{ICLP15}. This enables ILASP to perform a kind of preference
learning. For simplicity, we omitted details of weak constraint learning from
the main paper. In this section, we extend the concepts in the paper to show
how ILASP's new CDILP approach also extends to weak constraint learning.

\subsection*{Appendix A.1: Extra Background Material}
Unlike hard constraints in ASP, \emph{weak constraints} do not affect what is,
or is not, an answer set of a program $P$. Hence the above definitions also
apply to programs with weak constraints. Weak constraints create an ordering
over $AS(P)$ specifying which answer sets are ``better'' than others. A
\emph{weak constraint} is of the form $\asp{:\sim b_1,\ldots,b_n, \naf
c_1,\ldots,\naf c_m\ruleend[w@l,t_1,\ldots, t_k]}$ where $\asp{b_1, \ldots,
b_n,c_1,\ldots,c_m}$ are atoms, $\asp{w}$ and $\asp{l}$ are terms specifying
the \emph{weight} and the \emph{level}, and $\asp{t_1,\ldots,t_k}$ are terms
($\asp{[w@l,t_1,\ldots, t_k]}$ is called the \emph{tail} of the weak
constraint). At each \emph{priority level} $\asp{l}$, the aim is to discard any
answer set which does not minimise the sum of the weights of the ground weak
constraints (with level $\asp{l}$) whose bodies are true. The higher levels are
minimised first. Terms specify which ground weak constraints should be
considered unique.  For any program $P$ and $A\in AS(P)$, $\textit{weak}(P,A)$
is the set of tuples $\asp{(w,l,t_1,\ldots,t_k)}$ for which there is some
$\asp{:\sim b_1,\ldots,b_n,\naf c_1,\ldots,}$ $\asp{\naf
c_m\ruleend[w@l,t_1,\ldots,t_k]}$ in the grounding of $P$ such that $A$
satisfies $\asp{b_1,\ldots,b_n,}$ $\asp{\naf c_1,\ldots, \naf c_m}$.

In this section, the term ASP program is used to mean a program consisting of a
finite set of disjunctive rules and weak constraints.\footnote{The ILASP systems
support a wider range of ASP programs, including choice rules and conditional
literals, but we omit these concepts for simplicity.} The semantics of weak
constraints~\cite{ASPCORE2} are defined as follows.
%
  For each level $l$, $P^{l}_A = \sum_{(w,l,t_1,\ldots,t_k)\in \textit{\scriptsize weak}(P,
A)} w$.  For $A_1, A_2 \in AS(P)$, $A_1$ \emph{dominates} $A_2$ (written $A_1
\succ_{P} A_2$) iff $\exists l$ such that $P^{l}_{A_1} < P^{l}_{A_2}$ and
$\forall m > l, P^m_{A_1} = P^m_{A_2}$.  An answer set $A\!\in\! AS(P)$ is
\emph{optimal} if it is not dominated by any $A_2\!\in\!AS(P)$.

This paper uses the following notation to describe the preference relationship
between a pair of answer sets $A_1$ and $A_2$, using six binary comparison
operators $\lbrace <, >, \leq, \geq, =, \neq\rbrace$.

\begin{itemize}
  \item
    $A_1 <_{P} A_2$ iff $A_1 \succ_{P} A_2$ (i.e. iff $A_1$ dominates $A_2$);
  \item
    $A_1 >_{P} A_2$ iff $A_2 \succ_{P} A_1$ (i.e. iff $A_2$ dominates $A_1$);
  \item
    $A_1\leq_{P} A_2$ iff $A_2 \not\succ_{P} A_1$ (i.e. iff $A_2$ does not dominate $A_1$);
  \item
    $A_1\geq_{P} A_2$ iff $A_1 \not\succ_{P} A_2$ (i.e. iff $A_1$ does not dominate $A_2$);
  \item
    $A_1 =_P A_2$ iff $ A_1\not\succ_P A_2$ and $A_2 \not\succ_{P} A_1$ (i.e. iff neither $A_1$ nor $A_2$ dominates the other);
  \item
    $A_1\neq_{P} A_2$ iff $A_1 \succ_P A_2$ or $A_2 \succ_{P} A_1$ (i.e. iff either $A_1$ or $A_2$ dominates the other).
\end{itemize}

Let $WC$ be a set of weak constraints. The weak constraints in $WC$ are
\emph{independent} if there are no two weak constraints $W_1$ and $W_2$ in $WC$
with ground instances $W_1^g$ and $W_2^g$ such that the tail of $W_1^g$ is equal
to the tail of $W_2^g$. In this paper, we only consider programs whose weak
constraints are independent.

\paragraph{Ordering Examples.}
Positive and negative examples can be used to learn any ASP program consisting
of normal rules, choice rules and hard constraints.\footnote{This result holds,
up to strong equivalence, which means that given any such ASP program $P$, it
is possible to learn a program that is strongly equivalent to
$P$~\cite{AIJ17}.} As positive and negative examples can only express what
should or should not be an answer set of the learned program, they cannot be
used to learn weak constraints, which do not affect what is or is not an answer
set. Weak constraints create a preference ordering over the answer sets of a
program, so in order to learn them we need to give examples of this preference
ordering -- i.e.\ examples of which answer sets should be preferred to which
other answer sets.  These \emph{ordering examples} come in two forms:
\emph{brave orderings}, which express that at least one pair of accepting
answer sets for a pair of positive examples is ordered in a particular way; and
\emph{cautious orderings}, which express that every such pair of answer sets
should be ordered in that way.

A \emph{context-dependent ordering example} (CDOE) $o$ is a tuple $\langle e^1,
e^2, \prec\rangle$, where $e^1$ and $e^2$ are CDPIs and $\prec$ is a binary
comparison operator ($<$, $>$, $=$, $\leq$, $\geq$ or $\neq$).
A pair of interpretations $\langle I_1, I_2\rangle$ is said to be an
\emph{accepting pair of answer sets} of $o$ \wrt a program $P$ if all of the
following conditions hold: (i) $I_1 \in \textit{AAS}(e^1, P)$; (ii) $I_2\in
\textit{AAS}(e^2, P)$; and (iii) $I_1 \prec_{P} I_2$.
A program $P$ is said to \emph{bravely respect} $o$ if there is at least one
accepting pair of answer sets of $o$ \wrt $P$.
$P$ is said to \emph{cautiously respect} $o$ if there is no accepting pair of
answer sets of $\langle e^1, e^2, \prec^{-1}\rangle$ \wrt $P$ (where $<^{-1}$
is $\geq$, $>^{-1}$ is $\leq$, $\leq^{-1}$ is $>$, $\geq^{-1}$ is $<$, $=^{-1}$
is $\neq$ and $\neq^{-1}$ is $=$). In other words, $P$ bravely (resp.\
cautiously) respects $o$ if \emph{at least one} (resp.\ \emph{every}) pair of
answer sets extending the two CDPIs is ordered correctly (\wrt $\prec_{P}$).

\paragraph{Learning Task.} Definition~\ref{def:lnas} formalises the $\loasne$
\emph{learning task}, which is a generalisation of the $\lasne$ task used in
the main paper. A \emph{rule space} $S_{M}$ is a finite set of disjunctive
rules and weak constraints, defining the space of programs that are allowed to
be learned.

\begin{definition}\label{def:lnas}
  An $\loasne$ task $T$ is a tuple of the form $\langle B, S_{M}, \langle
  E^{+}, E^{-}, O^{b}, O^{c}\rangle\rangle$, where $B$ is an ASP program called
  the \emph{background knowledge}, $S_{M}$ is a rule space, $E^{+}$ and
  $E^{-}$ are (finite) sets of weighted CDPIs and $O^{b}$ and $O^{c}$ are
  (finite) sets of weighted CDOEs. Given a hypothesis $H \subseteq S_{M}$,

  \begin{enumerate}
    \item $\mathcal{U}(H, T)$ is the set consisting of: (a) all positive
      examples $e \in E^{+}$ such that $B \cup H$ does not accept $e$; (b) all
      negative examples $e \in E^{-}$ such that $B \cup H$ accepts $e$; (c) all
      brave ordering examples $o \in O^{b}$ such that $B \cup H$ does not
      bravely respect $o$; and (d) all cautious ordering examples $o \in O^{c}$
      such that $B \cup H$ does not cautiously respect $o$.
    \item
      the score of $H$, denoted as $\mathcal{S}(H, T)$, is the sum $|H| +
      \sum_{e \in \mathcal{U}(H, T)} e_{pen}$.
    \item $H$ is an inductive solution of $T$ (written $H \in \loasne(T)$) if
      and only if $\mathcal{S}(H, T)$ is finite.
    \item $H$ is an \emph{optimal inductive solution} of $T$ (written $H \in$
      $^*\loasne(T)$) if and only if $\mathcal{S}(H, T)$ is finite and
      $\nexists H' \subseteq S_{M}$ such that $\mathcal{S}(H, T) >
      \mathcal{S}(H', T)$.
  \end{enumerate}
\end{definition}

\subsection*{Appendix A.2: Conflict Analysis for CDOE's}
The notion of \emph{coverage constraint} is slightly extended for ordering
examples. This extension is formalised by the following definition.

\begin{definition}
  Let $S_M$ be a rule space. A \emph{coverage formula} over $S_M$ takes one of
  the following forms:
  \begin{itemize}
    \item $R_{id}$, for some $R\in S_M$.
    \item $\Sigma(w_1:R_{id}^1;\ldots;w_n:R_{id}^n) \prec w$, where
      $R_1,\ldots, R_n\in S_M$, $w_1,\ldots,w_n,w\in \mathbb{Z}$ and $\prec \in
      \lbrace <,>,\leq,\geq,=,\neq\rbrace$.
    \item $\lnot F$, where $F$ is a coverage formula over $S_M$.
    \item $F_1\lor\ldots\lor F_n$, where $F_1,\ldots,F_n$ are coverage formulas over $S_M$.
    \item $F_1\land\ldots\land F_n$, where $F_1,\ldots,F_n$ are coverage formulas over $S_M$.
  \end{itemize}

  The semantics of coverage formulas are defined as follows. Given a hypothesis $H$:

  \begin{itemize}
    \item $R_{id}$ accepts $H$ if and only if $R \in H$.
    \item $\Sigma(w_1:R_{id}^1;\ldots;w_n:R_{id}^n) \prec w$ accepts $H$ if and
      only if $\left(\sum\limits_{i\in \lbrack 1,n\rbrack, R^i \in H}w_i\right) \prec
      w$.
    \item $\lnot F$ accepts $H$ if and only if $F$ does not accept $H$.
    \item $F_1\lor\ldots\lor F_n$ accepts $H$ if and only if $\exists i\in [1,n]$ s.t. $F_i$ accepts $H$.
    \item $F_1\land\ldots\land F_n$ accepts $H$ if and only if $\forall i\in [1,n]$ s.t. $F_i$ accepts $H$.
  \end{itemize}

  A \emph{coverage constraint} is a pair $\langle e, F\rangle$, where $e$ is an
  example in $E$ and $F$ is a coverage formula, such that for any $H\subseteq
  S_M$, if $e$ is covered then $F$ accepts $H$.
\end{definition}

\subsubsection*{Conflict analysis for brave CDOE's}

Similarly to the iterative approach to conflict analysis for a positive CDPI
described in the main paper, ILASP uses an iterative approach to conflict
analysis for CDOE's.  Before formalising the method, it is necessary to
introduce some additional notation, and a way of computing a coverage formula
which can be used to determine whether a pair of interpretations are correctly
ordered.

\begin{definition}\label{def:opt_diff}
  Let $P$ be an ASP program and $I_1$ and $I_2$ be interpretations. For any
  integer $l$, the \emph{optimisation difference} between $I_1$ and $I_2$ at
  $l$ with respect to $P$ (denoted $\Delta^P_l(I_1, I_2)$) is equal to
  $P_l^{I_1} - P_l^{I_2}$.
\end{definition}

\begin{definition}\label{def:omegafrommeta}
  Let $A_1$ and $A_2$ be a pair of interpretations, $\prec$ be a binary
  comparison operator and $[l_1,\ldots,l_n]$ be the list of priority levels in
  $B\cup S_M$ (in descending order). For any $l\in[l_1,\ldots,l_n]$, $\omega(T,
  A_1, A_2, l, \prec)$ is the coverage formula $\Sigma(\Delta^{R^1}_l(A_1, A_2)
  : R_{id}^1;\ldots;\Delta^{R^n}_l(A_1, A_2) : R_{id}^n) \prec
  \Delta^{B}_l(A_1, A_2)$, where $\lbrace R^1,\ldots, R^n\rbrace = S_M$.

  Furthermore, $\omega(T, A_1, A_2, \prec)$ is the disjunction
  $F_1\lor\ldots\lor F_n$, where for each $i\in [1,n]$, $F_i = \omega(T, A_1,
  A_2, l_i, \prec) \land \omega(T, A_1, A_2, l_1, =)\land\ldots\land\omega(T,
  A_1, A_2, l_{i-1}, =)$.
\end{definition}

\appendixTheorem{omega}{
  Let $o$ be a CDOE, $\prec$ be a binary comparison operator and $A_1$ and
  $A_2$ be interpretations. A hypothesis $H\subseteq S_M$ satisfies $\omega(T,
  A_1, A_2, \prec)$ if and only if $A_1 \prec_{B \cup H} A_2$.
}{
  Let $H\subseteq S_M$.

  Assume that $H$ is accepted by $\omega(T, A_1, A_2, \prec)$.

  \proofindent{
    $\Leftrightarrow H$ is accepted by at least one disjunct $F_{i}$ in
    $\omega(T, A_1, A_2, \prec)$.

    $\Leftrightarrow \exists i\in [1,n]$ (where $n$ is as in
    Definition~\ref{def:omegafrommeta}) s.t.\ $\omega(T, A_1, A_2, l_i, \prec)
    \land \omega(T, A_1, A_2, l_1, =)\land\ldots\land\omega(T, A_1, A_2,
    l_{i-1}, =)$

    $\Leftrightarrow \exists i\in [1,n]$ s.t. $\Delta^{H}_{l_i}(A_1, A_2) \prec
    \Delta^{B}_{l_i}(A_1, A_2)$ and $\forall j\in [1,i-1]$,
    $\Delta^{H}_{l_i}(A_1, A_2) = \Delta^{B}_{l_i}(A_1, A_2)$.

    $\Leftrightarrow$ there is at least one priority level $l$ in $B\cup H$
    s.t.\ $(B\cup H)^{l}_{A_{1}} \prec (B\cup H)^{l}_{A_{2}}$ and for all
    higher priority levels $l'$ in $B\cup H$, $(B\cup H)^{l}_{A_{1}} = (B\cup
    H)^{l}_{A_{2}}$.

    $\Leftrightarrow$ $A_{1} \prec_{B\cup H} A_{2}$.
  }
}

\begin{algorithm}
  \begin{algorithmic}[1]
    \Procedure{iterative\_conflict\_analysis\_CDOE}{$\langle e_1, e_2, \prec\rangle, H, T, \psi$}
      \State{$F = \bot$;}
      \While{$\exists \langle I_1, I_2\rangle$ s.t.\ $\exists H\subseteq S_M$
      s.t.\ $F$ does not accept $H$, $I_1 \in \textit{AAS}(e_1, B\cup H)$, $I_2
      \in \textit{AAS}(e_2, B\cup H)$ and $I_{1} \prec_{B\cup H} I_{2}$}
        \State{Fix an arbitrary such $\langle I_1, I_2\rangle$}
        \State{$F = F \lor (\psi(I_1, I_2, e_1, e_2, \prec, H, T))$;}
      \EndWhile
      \State{\textbf{return}\,$F$;}
    \EndProcedure
  \end{algorithmic}
  \caption{$\textit{iterative\_conflict\_analysis\_CDOE}(\langle e_1, e_2, \prec\rangle, H, T, \psi)$ \label{alg:translateORD}}
\end{algorithm}

The $\textit{iterative\_conflict\_analysis\_CDOE}$ method relies on an extended
definition of the three $\psi$'s defined in the main paper. These are
formalised by the following definition.

\begin{definition}\label{def:psi_ord}
  Let $I$ be an interpretation.

  \begin{itemize}
    \item 
      If $I_1\in AAS(e_1, B\cup H)$ and $I_2\in AAS(e_2, B\cup H)$,
      $\psi_{\alpha}(I_1, I_2, e_1, e_2, \prec, H, T) = \omega(T, I_1, I_2, \prec)$.
      Otherwise, if $I_1$ and $I_2$ are both models of $H$, $\psi_{\alpha}(I_1,
      I_2, e_1, e_2, \prec, H, T)$ is an arbitrary conjunct of $\mathcal{T}_2(I_1,
      e_1, T)$ or $\mathcal{T}_2(I_2, e_2, T)$ that does not accept $H$.
      Otherwise, $\psi_{\alpha}(I_1, I_2, e_1, e_2, \prec, H, T) = \mathcal{T}_1(I_1,
      e_1, T)\land\mathcal{T}_1(I_2, e_2, T)$.
    \item 
      If $I_1\in AAS(e_1, B\cup H)$ and $I_2\in AAS(e_2, B\cup H)$,
      $\psi_{\beta}(I_1, I_2, e_1, e_2, \prec, H, T) = \omega(T, I_1, I_2, \prec)$.
      Otherwise, if $I_1$ and $I_2$ are both models of $H$, $\psi_{\beta}(I_1,
      I_2, e_1, e_2, \prec, H, T) = \mathcal{T}_2(I_1, e_1,
      T)\land\mathcal{T}_2(I_2, e_2, T)$ that does not accept $H$.  Otherwise,
      $\psi_{\beta}(I_1, I_2, e_1, e_2, \prec, H, T) = \mathcal{T}_1(I_1, e_1,
      T)\land\mathcal{T}_1(I_2, e_2, T)$.
    \item
      $\psi_{\gamma}(I_1, I_2, e_1, e_2, \prec, H, T) = \mathcal{T}(I_1, e_1,
      T)\land\mathcal{T}(I_2, e_2, T)\land\omega(T, I_1, I_2, \prec)$.
  \end{itemize}
\end{definition}

The following two theorems show that, for each of the three $\psi$'s, the\break
$\textit{iterative\_conflict\_analysis\_CDOE}$ procedure can be used to compute
coverage constraints for brave CDOEs. Specifically, they show that the method
is guaranteed to terminate and is a valid method for conflict analysis on brave
orderings.

\appendixTheorem{translateCDOEterminate}{
  Let $o = \langle e_1, e_2, \prec\rangle$ be a CDOE.
  For each $\psi \in \lbrace \psi_{\alpha}, \psi_{\beta},
  \psi_{\gamma}\rbrace$, the procedure\break
  $\textit{iterative\_conflict\_analysis\_CDOE}(o, H, T, \psi)$ is guaranteed
  to terminate and return a coverage formula $F_{\psi}$.  Furthermore, if $o$
  is a brave ordering, for each $\psi$ the pair $\langle o, F_{\psi}\rangle$ is
  a coverage constraint and there is no hypothesis accepted by $F_{\psi}$ that
  covers $o$.
}{
  First, we show that in each case, $\textit{iterative\_conflict\_analysis}(o,
  H, T, \psi)$ is guaranteed to terminate. As $F$ is a disjunction, and each
  iteration of the while loop adds an extra disjunct, the hypotheses
  $H'\subseteq S_M$ accepted must either increase or stay the same with each
  iteration. Hence as there are a finite number of hypotheses (because $S_M$ is
  finite), it remains to show that each iteration adds at least one hypothesis
  that is accepted by $F$ at the end of the iteration that was not accepted by
  $F$ at the start of the iteration.

  Next note, that in any iteration the hypothesis $H'$ referenced in the
  condition of the while loop is not accepted by $F$ at the start of the
  iteration. But as $I\in AAS(e, B\cup H')$, $H'$ must be accepted by
  $\mathcal{T}(I_1, e_1, T)$ and $\mathcal{T}(I_2, e_2, T)$ (by
  Theorem~\ref{thm:translateCDPI}) and by $\omega(T, I_1, I_2, \prec)$ (by
  Theorem~\ref{thm:translateCDOEterminate}); hence, as for each
  $\psi$ the formulas that are accepted by the conjunction of these three
  formulas are accepted by $\psi(I_1, I_2, e_1, e_2, \prec, H, T)$, $H'$ must be
  accepted by $F$ at the end of the iteration.
  Hence, in all three cases, $\textit{iterative\_conflict\_analysis}(o, H, T,
  \psi)$ is guaranteed to terminate.

  Next, we show that for each $\psi$, $\langle o, F_{\psi}\rangle$ is a
  coverage constraint. Assume for contradiction that it is not. Then there must
  be a hypothesis $H'$ that covers $o$ but is not accepted by $F_{\psi}$. This
  contradicts the termination of the while loop. Hence, $\langle o,
  F_{\psi}\rangle$ is a coverage constraint.

  Finally, we must show that $H$ is not accepted by $F_{\psi}$. If $H$ were
  accepted by $F_{\psi}$ then it must be accepted by one of the disjuncts of
  $F_{\psi}$, meaning that there must be an $I_1$ and $I_2$ such that $H$ is
  accepted by $\mathcal{T}(I_1, e_1, T)\land\mathcal{T}(I_2, e_2,
  T)\land\omega(T, I_1, I_2, \prec)$ (for each $\psi$, $\psi(I_1, I_2, e_1,
  e_2, \prec, H, T)$ is a consequence of $\mathcal{T}(I_1, e_1,
  T)\land\mathcal{T}(I_2, e_2, T)\land\omega(T, I_1, I_2, \prec)$). Hence, by
  Theorems~\ref{thm:translateCDPI} and~\ref{thm:translateCDOEterminate}, $I_1 \in AAS(B\cup
  H, e_1)$, $I_2 \in AAS(B\cup H, e_1)$ and $I_1 \prec_{B\cup H} I_2$. This
  contradicts the fact that $H$ does not cover $o$. Hence, $H$ is not accepted
  by $F_{\psi}$.
}

\subsubsection*{Conflict analysis for cautious CDOE's}

This section presents two methods of conflict analysis for a cautious CDOE $o$.
The first, used by ILASP3, is to call $iterative\_conflict\_analysis\_CDOE(o, H, T,
\psi_{\gamma})$. As the result of this is guaranteed to return a coverage
formula $F$ that is both necessary and sufficient for $e$ to be accepted, the
negation of this formula ($\lnot F$) is guaranteed to be necessary and
sufficient for $e$ to not be accepted -- i.e.\ for $e$ to be covered. This
result is formalised by Theorem~\ref{thm:neg_ord}.

\appendixTheorem{neg_ord}{
  Let $o=\langle e_1, e_2, \prec\rangle$ be a cautious CDOE (in $O^c$). Then\break
  $\textit{iterative\_conflict\_analysis\_CDOE}(\langle e_1, e_2,
  \prec^{-1}\rangle, H, T, \psi_{\gamma})$ is guaranteed to terminate,
  returning a coverage formula $F$.  Furthermore, the pair $\langle o, \lnot
  F\rangle$ is a coverage constraint and $H$ is not accepted by $\lnot F$.
}{
  Let $o_b$ be a brave ordering example $\langle e_1, e_2, \prec^{-1}\rangle$.

  By Theorem~\ref{thm:translateCDOEterminate},
  $\textit{iterative\_conflict\_analysis\_CDOE}(\langle e_1, e_2,
  \prec^{-1}\rangle, H, T, \psi_{\gamma})$ is guaranteed to terminate,
  returning a coverage formula $F$.

  It remains to show that the pair $\langle o, \lnot F\rangle$ is a coverage
  constraint and $H$ is not accepted by $\lnot F$.
  Assume for contradiction that there is a hypothesis $H'$ that covers $o$ that
  is not accepted by $\lnot F$. Then $H'$ must be accepted by one of the
  disjuncts of $F$. Hence, there must be an $I_1$ and $I_2$ such that $\langle
  I_1, I_2\rangle$ is an accepting pair of answer sets of $o_b$. This
  contradicts that $H'$ covers $o$. Hence, $\langle o, \lnot F\rangle$ is a
  coverage constraint.

  By Theorem~\ref{thm:translateCDOEterminate}, the pair $\langle o_b, F\rangle$
  is a coverage constraint, meaning that no hypothesis that covers $o_b$ can be
  accepted by $\lnot F$. As $H$ does not cover $o$, it must cover $o_b$, and
  hence, it is not accepted by $\lnot F$.
}

Similarly to the ILASP4 approach to conflict analysis for negative examples, a
coverage constraint can be computed for a cautious ordering example by finding
a single pair of interpretations that are ordered incorrectly by the current
hypothesis. For a hypothesis to cover the ordering example, it must either not
accept one of the two interpretations as an answer set, or it must order them
correctly. The coverage formula expressing these three possibilities is
formalised and proven to be correct by the following theorem. The computation
of this formula is guaranteed to terminate, and the theorem shows that using
this formula is a valid method for conflict analysis on cautious ordering
examples.

\appendixTheorem{cdpi_to_c_ord_nc}{
  Let $o = \langle e_1, e_2, \prec\rangle$ be a CDOE in $O^{c}$ and
  $H\subseteq S_{M}$ be a hypothesis that does not cover $o$. There is at least
  one pair of interpretations $\langle I_1, I_2\rangle$ such that $I_1 \in
  \textit{AAS}(e_1, B\cup H)$, $I_2 \in \textit{AAS}(e_2, B\cup H)$ and
  $I_1 \prec^{-1}_{B\cup H} I_{2}$:
  \begin{enumerate}
    \item $\lnot \mathcal{T}(I_1, e_1, T)\lor \lnot\mathcal{T}(I_2, e_2, T)
      \lor \omega(T,I_1, I_2, \prec)$ does not accept $H$.
    \item $\langle o, \lnot \mathcal{T}(I_1, e_1, T)\lor \lnot\mathcal{T}(I_2,
      e_2, T) \lor \omega(T,I_1, I_2, \prec)\rangle$ is a coverage constraint.
  \end{enumerate}
}{
  As $H$ does not cover $o$, there must be at least one accepting pair of
  answer sets of $\langle e_1, e_2, \prec^{-1}\rangle$ \wrt $B\cup H$ (by definition
  of a cautious ordering example).
  \begin{enumerate}
    \item Let $\langle I_1, I_2\rangle$ be an arbitrary pair of interpretations
      $\langle I_1, I_2\rangle$ such that $I_1 \in \textit{AAS}(e_1, B\cup H)$,
      $I_2 \in \textit{AAS}(e_2, B\cup H)$ and $I_1 \prec^{-1}_{B\cup H}
      I_{2}$. By Theorem~\ref{thm:translateCDPI}, $\mathcal{T}(I_1, e_1, T)$
      and $\mathcal{T}(I_2, e_2, T)$ both accept $H$ and by
      Theorem~\ref{thm:omega} $\omega(T,I_1, I_2, \prec)$ does not accept $H$.
      Hence, $\lnot \mathcal{T}(I_1, e_1, T)\lor \lnot\mathcal{T}(I_2, e_2, T)
      \lor \omega(T,I_1, I_2, \prec)$ does not accept $H$.
    \item
      Let $H'$ be a hypothesis that covers $o$. It suffices to show that $H'$
      is accepted by $\lnot \mathcal{T}(I_1, e_1, T)\lor \lnot\mathcal{T}(I_2,
      e_2, T) \lor \omega(T,I_1, I_2, \prec)\rangle$. As $H'$ covers $o$,
      $\langle I_1, I_2\rangle$ cannot be an accepting pair of answer sets of
      $\langle e_1, e_2, \prec^{-1}\rangle$ \wrt $B\cup H'$. Hence, either $I_1
      \not\in \textit{AAS}(e_1, B\cup H)$, $I_2 \not\in \textit{AAS}(e_2, B\cup
      H)$ or $I_1 \prec_{B\cup H} I_{2}$; and hence, (by
      Theorems~\ref{thm:translateCDPI} and~\ref{thm:omega}) $H'$ is accepted by
      $\lnot \mathcal{T}(I_1, e_1, T)\lor \lnot\mathcal{T}(I_2,
      e_2, T) \lor \omega(T,I_1, I_2, \prec)\rangle$.
  \end{enumerate}
}

\subsection*{Appendix A.3: Evaluation of CDILP on Preference Learning Tasks}
This section presents some extra benchmarks used to evaluate the CDILP approach
on weak constraint learning tasks.

\begin{table}
  \footnotesize
  \begin{tabular}{c ccccc ccccc}
    \hline
    Task   & $|S_M|$ & $|E^{+}|$ & $|E^{-}|$ & $|O^{b}|$ & $|O^{c}|$ & 2i & 3 & 4a & 4b \\\hline\hline
    Scheduling (3 day)      & 180       & 400 & 0   & 110 & 90  & 10.14 & 17.01 & 3.83  & 4.25 \\
    Scheduling (4 day)      & 180       & 400 & 0   & 128 & 72  & 29.04 & 36.16 & 5.17  & 5.15 \\
    Scheduling (5 day)      & 180       & 400 & 0   & 133 & 67  & 53.84 & 22.63 & 6.55  & 6.81 \\\hline
    Agent D                 & 244       & 172 & 228 & 390 & 0   & 84.51 & 63.38 & 38.63 & 34.99\\\hline
    Journey                 & 117       & 386 & 0   & 200 & 0   & 1.35  & 2.92  & 1.54 & 1.66 \\
  \end{tabular}
  \caption{The running times (in seconds) of various ILASP systems on the set
  of benchmark problems. TO denotes a timeout (where the time limit was
  1800s).\label{tbl:benchmarks}}
\end{table}

The first additional benchmark problem is that of learning scheduling
preferences, first presented in~\cite{ICLP15}. In this setting, the goal is to
learn an academic's preferences about interview scheduling, encoded as weak
constraints. The tasks in this case are over examples with \texttt{3x3},
\texttt{4x3} and \texttt{5x3} timetables, respectively (i.e.\ three day, four
day and five day timetables).  In this case, ILASP2i and ILASP3 perform fairly
similarly, but both versions of ILASP4 are significantly better either ILASP2i
or 3. This task is non-categorical, but in fact as there are only weak
constraints in the hypothesis space, for brave orderings ILASP3, ILASP4a and
ILASP4b are actually guaranteed to compute the same coverage formula. However,
for cautious orderings this is not the case, and the two ILASP4 algorithms will
tend to compute shorter formulas than ILASP3 (corresponding to a single pair of
answer sets that were ordered incorrectly by the previous hypothesis, rather
than all possible pairs of answer sets), although there is still no difference
between the two ILASP4 algorithms in this case. This explains the improvement
in performance of ILASP4 over ILASP3.

The Agent D learning task is an extension of the Agent benchmarks used in the main paper.
In addition to the rules and hard constraints learned in Agent C, weak
constraints must be learned to explain why some traces through the grid are
preferred to others.  This uses positive, negative and brave ordering examples.

The final benchmark is based on a dataset from~\cite{ICLP16}, in which the goal
is to learn a user's journey preferences from examples of which journeys the
user prefers over other journeys. This task is categorical and contains only
weak constraints. For such tasks, ILASP3, ILASP4a and ILASP4b will compute the
same coverage formulas in all cases, so there is not much difference between
them -- minor details of the different implementations cause the ILASP4
approaches to still be slightly faster. All approaches perform similarly to
ILASP2i on this task.

%

\end{document}